\documentclass[twoside,11pt]{article}

\usepackage{booktabs}
\usepackage{multirow}
\usepackage{hhline}
\usepackage{tabularx}
\usepackage{amsmath}
\usepackage{subcaption}
\usepackage{graphicx}
\usepackage{tikz}
\usepackage[skip=10pt]{caption}
\usetikzlibrary{shapes, arrows.meta, positioning, calc, fit}
\usepackage{makecell} 

\usepackage[preprint]{dmlr2e}

\ShortHeadings{ASD-FEAT}{Liaqat et al.}
\firstpageno{1}

\begin{document}

\title{ASD-FEAT: A Multi-Modal Infant Video-Derived Dataset for Early ASD Risk Prediction}

\author{\name Sidrah Liaqat,
        \name Halil Helvaci \and
        \name Sen-Ching Cheung \\
       \addr Department of Electrical and Computer Engineering\\
       University of Kentucky\\
       Lexington, KY 40506, USA
       \email sccheung@ieee.org
       \AND
       \name Chongruo Wu,
       \name Dongjie Chen \and
       \name Chen Nee Chuah \\
       \addr Department of Electrical and Computer Engineering \\
       University of California\\
       Davis, CA 95616, USA
       \AND
       \name Sally Ozonoff \\
       \addr Department of Psychiatry and Behavioral Sciences\\
       MIND Institute \\
       University of California\\
       Davis, CA 95616, USA
       }

\maketitle

\begin{abstract}

Accurate early screening for Autism Spectrum Disorder (ASD) is a precursor to timely intervention, which is critical for improving cognitive and behavioral outcomes. We present ASD-FEAT (ASD - Feature Extraction And Tracking), a multimodal dataset derived from video recordings of infant--adult interaction sessions. The dataset is designed to support ASD behavior recognition for early screening research. The key contribution of ASD-FEAT is the combination of longitudinal coverage from infancy through 36 months, repeated interaction sessions, clinically validated developmental outcomes, expert frame-level behavioral annotations, and privacy-conscious multimodal feature representations. To the best of our knowledge, existing ASD behavioral datasets do not jointly provide these characteristics at comparable scale. To demonstrate the utility of ASD-FEAT, we use it to evaluate a computer-vision-based end-to-end pipeline relying on machine learning techniques to automatically identify ASD risk. ASD-FEAT integrates both expert-defined and deep-learned features, including face and eye landmarks, facial action units, gaze direction, head position, mel-spectrogram audio representations, and optical flow, to identify behavioral markers of social interaction. Our automated pipeline achieves an ASD classification accuracy of 76.2\% and an Area Under the Receiver Operating Characteristic (AUROC) of 0.82, compared to classifiers trained on manually labeled behaviors, which yielded 81.3\% accuracy and an AUROC of 0.88. We further introduce a within-visit partner contrast: a per-visit signal contrasting examiner-directed and parent-directed social behavior which, when added to the classifier, lifts the fully automated Look Face~+~Smile and Look Face~+~Vocal configurations to Matthews correlations of 0.49 and 0.45 respectively, exceeding the human-coded single-partner baseline of 0.42. These results highlight the feasibility of automated behavioral analysis in ASD screening and provide a baseline for future research.
\end{abstract}


\section{Introduction}

Autism Spectrum Disorder (ASD) is a developmental disorder affecting 1 out of 31 (3.2\%) children in the U.S. ~\citep{shaw2025prevalence}. ASD presents as deficits in social and communicative development, as well as stereotyped or restricted behaviors and interests. During social interactions, infants and toddlers with ASD make less eye contact and smile at others less than typically developing (TD) children, and are often more interested in objects than people. The median age of ASD diagnosis in the United States is still over 4 years \citep{shaw2025prevalence} resulting in missed opportunity for intervention during a critical developmental window when treatment is most effective in improving autistic symptoms and behaviors \citep{guthrie2023earlier}.

Early interventions have been shown to yield positive outcomes for children diagnosed with ASD \citep{zwaigenbaum2015early}.  Accurate and widespread early diagnosis of ASD, which is a critical prerequisite to initiate treatment, is important  for optimizing treatment outcomes for children at high risk of ASD. Timely access to diagnostic examination is challenging due to long wait times and limited availability of clinicians trained in autism diagnosis ~\citep{chen2023delay, malik2022tackling}. In addition, some ASD symptoms are subtle or may be overlooked, particularly during the early years of a child's development ~\citep{dawson2023prediction}. These two challenges in early screening may lead to delayed treatment and missed opportunity for improved treatment outcomes. 

Recognition of deficits in social-emotional reciprocity and nonverbal communication is central to early identification of ASD \citep{DSM5, zwaigenbaum2015early}. By analyzing multimodal behavioral data such as facial expressions, eye gaze, vocal prosody, and gesture patterns, machine learning models can be trained to detect subtle atypicality that may elude human observation \citep{saakyan2025improving,malik2022tackling,rakotomanana2025scoping}. These automated approaches have the potential of not only enhancing diagnostic accuracy but also enabling unbiased, scalable, low-cost early screening, particularly in underserved or resource-limited settings \citep{dubey2024mobile, mukherjee2024digital, marlow2019review}. Recent work in digital phenotyping has demonstrated promising results for automated ASD screening \citep{perochon2023early}. However, many existing approaches rely on controlled laboratory environments with specialized equipment \citep{hashemi2018computer} or pre-selected stimuli, limiting their real-world applicability. Moreover, existing datasets for training ASD-related behavior classifiers are often limited in size, lack longitudinal data, or have inconsistent annotation quality that limits their utility in developing robust, generalizable models (see Section \ref{existing_behavioral_datasets} for a detailed discussion). To address these limitations and advance the development of accessible, naturalistic ASD screening tools, there is a critical need for large-scale, longitudinally collected datasets with high-quality behavioral annotations captured in real-world settings.

\begin{table*}[t]
\footnotesize
\centering
\caption{Comparative analysis of multimodal datasets for pediatric ASD research, detailing sample sizes, video durations, behavioral annotations, and diagnostic assessment methods.}
\label{tab:dataset_comparison}

\begin{tabular}{lllllll}
\toprule
\textbf{Dataset} & \makecell[l]{\textbf{No. of}\\\textbf{Subjects}} & \makecell[l]{\textbf{Video}\\\textbf{Source}} & \makecell[l]{\textbf{Ages}\\\textbf{(m)}} & \makecell[l]{\textbf{Video}\\\textbf{Duration}} & \makecell[l]{\textbf{Annotated}\\\textbf{Behaviors}} & \makecell[l]{\textbf{ASD}\\\textbf{Diagnostic}\\\textbf{Method}} \\ 
\midrule
\makecell[l]{MMDB \\ \citep{rehg2013decoding}} & 121 & Lab & 15--30 & \makecell[l]{160 videos\\(5 min each)} & \makecell[l]{Gaze, facial exp, \\ gestures, vocal.} & \makecell[l]{Rapid\\ABC} \\
\addlinespace 
\makecell[l]{DE-ENIGMA \\ \citep{riva2020enigma}} & \makecell[l]{50\\(25 ASD)} & Lab & 60--144 & \makecell[l]{$\sim$250 videos\\(15--20 min)} & \makecell[l]{Affect, body\\gestures, vocal.} & ADOS-2 \\
\addlinespace
\makecell[l]{ChildPlay Gaze \\ \citep{tafasca2023childplay}} & \makecell[l]{95 videos\\(401 clips)} & YouTube & 36--144 & \makecell[l]{$\sim$67 min\\(total\\duration)} & Gaze point/class & -- \\
\addlinespace
\makecell[l]{MMASD \\ \citep{li2023mmasd}} & 32 ASD & Interv. & -- & \makecell[l]{1315 videos\\(7 sec each)} & 11 Body actions & ADOS-2 \\
\addlinespace
\textbf{ASD-FEAT (ours)} & \makecell[l]{318\\(54 ASD)} & Lab & 6--36 & \makecell[l]{1886 videos\\(3 min each)} & \makecell[l]{Gaze to face/obj, \\ smile, vocal.} & ADOS-2 \\
\bottomrule
\end{tabular}
\end{table*}

We introduce a feature dataset derived from video, ASD-FEAT, specially designed to support the development of models for analyzing behaviors relevant to ASD risk prediction. This dataset captures natural dyadic interactions between adults and infants/toddlers in naturalistic, semi-clinical settings. The underlying videos feature longitudinal recordings of the same children over a span of 30 months, allowing for the study of developmental trajectories. 
In contrast to prior ASD-related datasets such as MMDB~\citep{rehg2013decoding}, DE-ENIGMA~\citep{riva2020enigma}, ChildPlay Gaze~\citep{tafasca2023childplay}, and MMASD~\citep{li2023mmasd} (see Table~\ref{tab:dataset_comparison}), which are limited in size, scope, or annotation quality, our ASD-FEAT dataset provides a significantly larger and professionally coded resource with frame-level annotations for clinically relevant behaviors. Our dataset captures the same subjects longitudinally across multiple time points in early development, each assessed through gold-standard ADOS-2 protocols and DSM-based clinical diagnoses. To better protect participant privacy, raw video is not released; instead it is transformed into derived feature representations rather than shared directly. This combination of scale, diagnostic rigor, temporal depth, and privacy-aware design makes ASD-FEAT particularly well suited for developing and benchmarking models for early ASD behavioral analysis and risk prediction.

\begin{figure*}[t]
    \centering
    \includegraphics[width=0.80\textwidth]{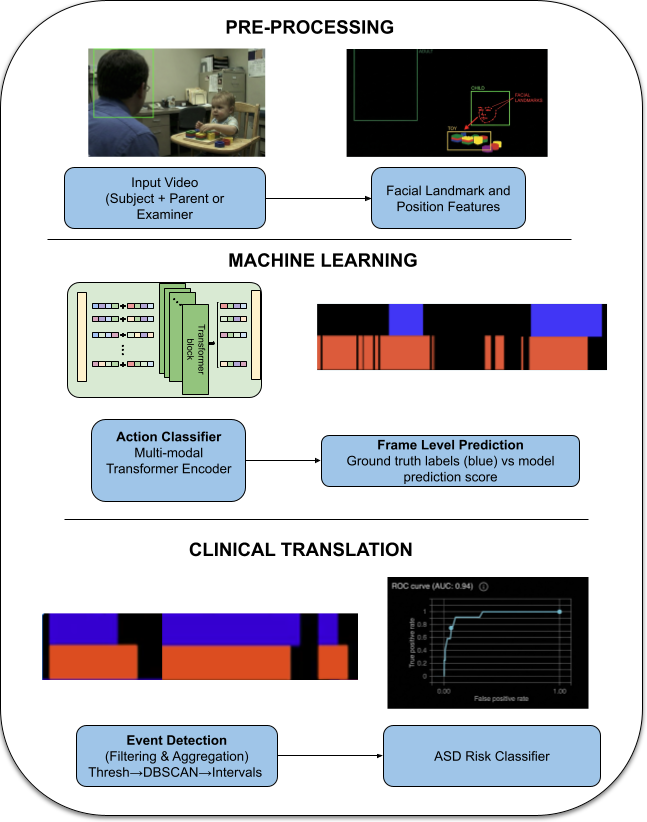} 
    \caption{Computer-vision-based ASD risk prediction pipeline. Following pre-processing, the machine learning stage detects fine-grained ASD-related behaviors using features derived from clinical interaction videos. The last stage predicts ASD risk based on manually annotated or machine-predicted behavioral markers. To protect subject privacy, the images in this figure are generated examples rather than frames from actual clinical videos.}
    \label{fig:pipeline}
\end{figure*}

In addition, we propose a baseline machine learning methodology for ASD-related behavior recognition (Figure~\ref{fig:pipeline}) that  complements end-to-end appearance-based models by leveraging a set of interpretable, behavior-specific features tailored for key social behaviors. These features include face and eye landmarks, head position, eye gaze direction, facial action units (AUs), mel-spectrograms, and optical flow. While such expert-derived features may be considered less flexible than deep-learning representations, they offer significant advantages in interpretability, ease of deployment in real-world settings without specialized hardware, and the ability to operate without compromising data privacy. Previous studies have already identified the usefulness of many of these features in detecting behaviors relevant to ASD~\citep{jaiswal2017automatic}. Moreover, these features are semantically and clinically meaningful, providing behavioral context and enhancing interpretability ~\citep{murdoch2019interpretable} of the classification model's performance. Our approach allows for accurate detection of social behaviors such as mutual social gaze, social smiles and gaze towards objects, without the need for complex camera configurations, calibration, depth sensing, wearable devices, or eye tracking, making it highly suitable for early screening scenarios in naturalistic environments. 

Our contributions in this work are as follows:
\begin{enumerate}
    \item We release a de-identified feature dataset for public use comprising face and eye landmarks, facial AUs, head location, eye gaze direction, mel-spectrogram audio features as well as deep-learned color and optical flow features. The dataset also includes frame-level behavioral annotations performed by expert coders trained to rigorous reliability standards. 
    \item We introduce a practical approach for fine-grained visual attention detection in social interaction videos using transformer-based architecture with emphasis on interpretable, behavior-specific features.
    \item Our proposed method leverages facial and eye landmark features, along with head and object position information, to identify gaze behavior at the frame level, designed to accommodate non-frontal head poses and occlusions commonly encountered in naturalistic interaction videos. Baseline classification models for positive affect and vocalization detection are also presented. 
    \item We present a computer vision-assisted ASD risk prediction framework that analyzes key ASD-related behaviors. Using automatically extracted features, the framework achieves an ASD classification accuracy of 76.2\% and an AUROC score of 0.82, compared to classification using manually labeled behaviors with 81.3\% accuracy and 0.88 AUROC score. These results highlight the feasibility of automated behavioral analysis in ASD screening and provide a baseline for future research.
    \item Beyond aggregate behavioral rates, we introduce a within-visit partner contrast (the contrast between a child's behavior with the examiner and with the parent) which, when integrated into the Stage~2 classifier, drives fully automated screening based on gaze plus positive affect and gaze plus vocalization signals to Matthews correlations of 0.49 and 0.45 respectively, surpassing the human-coded single-partner baseline of 0.42.
\end{enumerate}

\section{Related Work}

Automated early screening for ASD has become increasingly explored with advances in computer vision and machine learning in recent years~\citep{perochon2023early, saakyan2025improving, mukherjee2024digital}. In this section, we first review traditional clinical practices for ASD screening and diagnosis, highlighting barriers to their widespread implementation. We then survey machine learning approaches for ASD risk prediction, organized by data modalities. Finally, we discuss existing ASD-related datasets and identify key limitations in current automated video analysis approaches that motivate our work.

\subsection{Clinical Practice in ASD Screening and Diagnosis}
\label{sec:clinical_screening_diagnosis}
The gold standard for ASD diagnosis involves a comprehensive clinical assessment conducted by licensed professionals using standardized instruments such as the Autism Diagnostic Interview–Revised (ADI-R) \citep{lord1994autism} and the Autism Diagnostic Observation Schedule (ADOS) \citep{lord2012autism}. However, access to such evaluation is often limited by long wait times and a shortage of trained specialists. A large national health-research-network study found an average delay of over two years between a child's initial developmental screening and their ASD diagnosis~\citep{chen2023delay}. Such extensive delays between screening and diagnosis serve as a strong catalyst for this research, demonstrating a clear need for scalable, computational tools that can augment traditional diagnostic workflows.

In contrast, screening involves the use of brief, accessible tools, often used as part of routine well-child checkups, designed to flag potential concerns and identify children in need of a comprehensive evaluation for ASD. Among the most widely used parent-completed screening questionnaires is the Modified Checklist for Autism in Toddlers, Revised, with Follow-up (M-CHAT-R/F) \citep{robins2014validation}, designed to assess risk in children aged 16 to 30 months by evaluating behaviors related to social interaction, communication, and repetitive behaviors and interests. Another important tool, often used for even younger children (6-24 months) and part of the Communication and Symbolic Behavior Scales (CSBS DP™) \citep{wetherby2003early}, is the Infant-Toddler Checklist (ITC) \citep{ITCwetherby2008validation}, which focuses on early communication and symbolic abilities. 

Despite the availability of validated ASD screening tools with reported sensitivities ranging from 85\% to 91\%~\citep{robins2014validation}, their consistent implementation is met with multiple barriers. At the provider level, clinicians contend with time constraints during visits, insufficient training in screening interpretation, and poor integration of screening workflows into electronic health records~\citep{hightshoe2023barriers}. Caregiver-level barriers include language differences, low health literacy, cultural variations in understanding social behaviors, and concerns about stigma that can impact screening accuracy and participation~\citep{khowaja2015sociodemographic}. At the systems level, inadequate reimbursement, lengthy diagnostic wait times, and limited referral infrastructure frequently discourage routine screening~\citep{chen2023delay}.

Collectively, these systemic barriers highlight a growing need for scalable alternatives that can augment clinical capacity while maintaining diagnostic fidelity. To bridge the gap between population-level screening demands and the necessity for individualized care, this work proposes a novel video-based dataset and computational framework designed to enable precision phenotyping of ASD-related behaviors. By leveraging computer vision and machine learning to extract multimodal behavioral signatures from naturalistic video recordings, our proposed approach moves towards nuanced, objective characterization of social-communicative phenotypes of ASD-related behaviors.

\subsection{Behavioral Biomarkers for ASD Detection}

Machine learning approaches to ASD prediction using behavioral biomarkers have emerged as a promising pathway for scalable, non-invasive early screening. Computational analysis of video-recorded interactions enables more objective, quantifiable measurement of subtle behavioral differences. These automated approaches are particularly valuable for addressing the screening barriers identified in Section \ref{sec:clinical_screening_diagnosis}, as they can be deployed in naturalistic settings without specialized equipment or extensive clinical expertise.

Behavioral biomarkers for ASD span multiple domains, each capturing distinct aspects of social-communicative development. We organize our review by modality, examining how computer vision and machine learning have been applied to analyze eye gaze, motor patterns, facial expressions, and vocalizations, before discussing multimodal integration approaches.

\subsubsection{Eye Gaze and Visual Attention}

Different approaches to gaze detection involve distinct tradeoffs. Hardware-based eye trackers provide high accuracy but typically require controlled clinical or laboratory settings, standardized viewing distances, camera calibration, and frontal face orientation \citep{mayrand2023dual, bradshaw2023infant}. Appearance-based deep learning methods can handle more varied poses but require extensive training data and high computational resources \citep{marin2019laeo,chong2020detecting,tafasca2023childplay}. Feature-based approaches using facial landmarks and 3D gaze estimation offer a middle ground, providing improved robustness to non-frontal poses with lower data requirements.

Beyond the technical challenge of tracking, the target of visual attention is a critical differentiator. Research has consistently shown that autistic children demonstrate reduced relative interest in social stimuli (faces, eyes) compared to non-social stimuli (objects, natural scenes) \citep{Duan18, shic2022autism}. Consequently, automated systems must not only estimate gaze direction but also contextualize it within the scene to quantify attention toward social partners versus background objects.

In \cite{guo2022mgtr}, transformer-based architectures were applied to model temporal dependencies in gaze behavior. The method was benchmarked on the UCO-LAEO and AVA-LAEO datasets, where it achieved mean average precision (mAP) scores of 64.8 and 66.2 on the mutual gaze detection task. However, most existing methods still struggle with the highly naturalistic conditions common in clinical recordings. Unlike controlled benchmark datasets, clinical interactions often capture the child in semi-profile and the adult clinician from the back of the head, which can limit the utility of off-the-shelf frontal face detectors. Furthermore, children frequently look away or move rapidly, leading to motion blur and partial occlusions that disrupt continuous tracking. These challenges motivate our use of 3D gaze estimation and contextual cues (head and object positions) to enable robust gaze detection in non-frontal, partially occluded scenarios.

\subsubsection{Facial Expressions and Affect}

Facial expression analysis has focused on both the presence and temporal dynamics of social affect. Children with ASD exhibit reduced frequency and intensity of social smiles—those directed at a communication partner in response to social interaction—compared to typically developing ones \citep{ozonoff2010prospective}. These differences emerge early, with reduced social smiling observable by 12 months in infants later diagnosed with ASD \citep{Ozonoff2019}.

Computational approaches leverage multiple levels of facial analysis. Facial Action Unit (AU) detection systems based on the Facial Action Coding System quantify specific muscle movements (e.g., AU12 for lip corner pulling in smiles), enabling fine-grained analysis of expression quality and timing \citep{baltrusaitis2018openface}. Appearance-based deep learning methods trained on emotion datasets can classify expressions but may miss subtle ASD-related differences. Recent work has shown that analyzing facial landmark geometry combined with temporal dynamics can better capture these nuances, achieving 87.8\% sensitivity in identifying ASD-related affect patterns \citep{perochon2023early}.

A key challenge is that facial expressions in ASD are often present but exhibit subtle differences in timing, intensity, and context-appropriateness. This makes detection more nuanced than simple presence/absence classification and motivates the use of frame-level modeling with attention to temporal context. We leverage facial landmarks and AU-based features from OpenFace rather than end-to-end appearance learning to ensure interpretability while maintaining performance with limited training data.

\subsubsection{Vocal and Prosodic Features}

Vocal and prosodic characteristics provide a complementary modality that directly captures communication behaviors. Children with ASD show reduced frequency of social vocalizations compared to typically developing children \citep{oller2024acoustic}. Beyond frequency, prosodic differences, including atypical pitch contours, reduced intonation variability, and unusual rhythm, have been documented in both verbal and pre-verbal children with ASD \citep{weed2024different}.

Automated vocal analysis has employed diverse approaches. Traditional audio processing extracts handcrafted prosodic features such as pitch (F0), formants, jitter, and shimmer for classification \citep{oller2024acoustic}. More recently, deep learning methods operating on Mel frequency cepstral coefficients (MFCC) have achieved up to 98\% accuracy in detecting atypical vocalizations \citep{fagherazzi2021voice}. The LENA system  \citep{putnam2025use} automates long-form home recording analysis, quantifying overall vocal production and conversational turn-taking, though it often requires manual validation for fine-grained behavior coding.

Challenges in automated vocal analysis include distinguishing child vocalizations from background speech, handling noisy home environments, and the high variability in typical vocal development across age. Additionally, many existing audio-based ASD detection systems are appropriate primarily for older verbal children, with fewer validated approaches for pre-verbal infants where early screening is most critical. Mel spectrogram-based representations offer a practical balance between discriminative power and computational efficiency for frame-level vocalization detection.

\subsubsection{Motor Movement Patterns}

While our automated ASD screener focuses on social behaviors, motor patterns provide another important behavioral biomarker. Motor differences are increasingly recognized to be among the core features of ASD. Children with ASD often exhibit atypical motor development including delayed motor milestones, reduced postural control, and stereotyped repetitive movements such as hand flapping and body rocking \citep{zampella2021motor,bhat2021motor}.

Computer vision approaches to motor analysis have taken several forms. Pose estimation methods track body joint positions over time to quantify postural stability and movement smoothness \citep{zheng2023deep}. Touchscreen-based gameplay has been leveraged to capture fine motor patterns, with machine learning identifying ASD using ``motor signatures" from touch dynamics with up to 93\% accuracy \citep{MOTORanzulewicz2016toward}. More recently, deep learning applied to video has enabled automated detection of 92.5\% of stereotyped motor behaviors with 66.8\% Positive Predictive Value (PPV) \citep{barami2024automated}.

Although approaches to quantifying motor patterns, ranging from pose estimation to touchscreen dynamics, represent a significant area of ASD research, they address a behavioral domain distinct from the social-communicative deficits central to this study. However, given the challenges of variability and the need for full-body constraints in motor analysis, our work specifically prioritizes the detection of interpersonal social signals which serve as more direct indicators of shared engagement in dyadic interactions. 

\subsubsection{Multimodal Integration}

Given the heterogeneity of ASD presentation and the complementary nature of different behavioral domains, multimodal approaches that integrate multiple biomarkers have shown substantial advantages over single-modality methods. A digital phenotyping app called SenseToKnow exemplifies this paradigm by combining head posture control, gaze tracking, facial expressions, motor behaviors, and blink rate in a structured preferential looking task, achieving 87.8\% sensitivity and 80.8\% specificity \citep{perochon2023early}.

Recent work has demonstrated the benefits of multimodal fusion for naturalistic video analysis. \citep{saakyan2025improving} combined facial expressions, voice prosody, head motion, heart rate variability, and gaze behavior using late fusion, achieving 74\% classification accuracy; an improvement over the best single-modality result of 69\% from gaze alone. Similarly, comprehensive reviews of AI-based approaches highlight that systems integrating voice, facial, and behavioral phenotyping consistently outperform unimodal methods \citep{saakyan2025improving,rakotomanana2025scoping}.

Multimodal fusion strategies vary in complexity and effectiveness. Early fusion, which concatenates features from different modalities before learning, is simple but may dilute discriminative signals. Late fusion, which trains separate models per modality and combines predictions, can better handle modality-specific patterns and missing data. Attention-based fusion mechanisms that learn to weight modalities dynamically show promise but require larger datasets.

Key challenges in multimodal approaches include temporal alignment of different data streams (e.g., audio and video), handling missing modalities in some recordings, increased computational requirements, and the need for larger annotated datasets. Despite these challenges, multimodal methods represent the current state-of-the-art for behavioral ASD prediction. Motivated by these findings, our framework integrates visual and audio features through a transformer-based architecture that can model temporal dependencies across modalities.

\subsection{ Existing ASD Behavioral Datasets}
\label{existing_behavioral_datasets}
The development of robust machine learning models for ASD screening depends critically on high-quality, well-annotated datasets. The key characteristics of the major datasets used in the literature are summarized in the following subsections.
\subsubsection{MMDB (Multimodal Dyadic Behavior Dataset)}
MMDB~\citep{rehg2013decoding} was an early resource for analyzing social behaviors, capturing semi-structured play sessions of children aged 1-2 years. While it established early benchmarks for gaze and engagement analysis~\citep{chong2020detecting}, it is relatively small (160 sessions from 121 children) and lacks the comprehensive frame-level annotations required for modern deep learning. 
\subsubsection{DE-ENIGMA Dataset}

The DE-ENIGMA dataset~\citep{riva2020enigma} provides multimodal behavioral data including audio, video, and RGB-D motion tracking recorded during human-robot interactions across multiple European sites. While the inclusion of standardized clinical measures supports data consistency, the structured, robot-mediated environment may limit the ecological validity of the behaviors captured compared to naturalistic, child-led settings. Furthermore, access to the full 150+ hours of data is restricted to academic researchers under specific licensing agreements ~\citep{mukherjee2024digital}. Of these 150+ hours, only a smaller subset carries behavioral annotations: 50 continuously labeled session recordings (one per participant---25 with ASD and 25 without---roughly 15--20 minutes each), each captured from five synchronized viewpoints (three RGB cameras and two Microsoft Kinect RGB-D sensors) and thus comprising approximately 250 parallel multi-view video streams. These sessions were coded frame-by-frame by trained expert coders for continuous affect (valence and arousal), body gestures and posture (3D skeleton joints, spatial movement, and posture metrics), and vocalizations (acoustic features and child--therapist--robot speaker diarization). The dataset's focus on school-aged children (5--12 years) offers limited utility for investigating the infant and toddler period markers (0--36 months) currently identified as critical for early-staged screening and intervention ~\citep{dawson2023prediction}.

\subsubsection{ChildPlay Gaze Dataset}
ChildPlay Gaze~\citep{tafasca2023childplay} provides high-quality manual gaze-target annotations for children in naturalistic, uncontrolled settings, with clips curated from in-the-wild online videos rather than collected under a controlled clinical protocol. While this makes it a valuable benchmark for gaze-following models in young children, the clips are not drawn from a screening context and carry no diagnostic outcomes, and the dataset is restricted to gaze, lacking other key modalities such as facial expression or vocalization~\citep{rakotomanana2025scoping}.
\subsubsection{MMASD (Multimodal ASD Dataset)}
MMASD~\citep{li2023mmasd} advances the field by capturing synchronized video and audio of therapy sessions. While valuable for analyzing intervention strategies, the participants are already diagnosed and receiving treatment, limiting the dataset’s utility for training screening tools meant to distinguish undiagnosed at-risk children. The annotations also focus on therapy-specific behaviors rather than the subtle, frame-level social cues (e.g., fleeting eye contact) used in early detection.
\subsubsection{Other Relevant Datasets}
Beyond these major datasets, several other resources have contributed to ASD research, though each has specific scope or access limitations. The Autism Diagnostic Observation Schedule (ADOS) video datasets maintained at various research institutions contain rich clinical assessment data but are rarely publicly available due to privacy concerns and institutional restrictions~\citep{ahn2024objective, lord2012autism}.
Mobile health and digital phenotyping studies have generated proprietary datasets such as those from the SenseToKnow app~\citep{perochon2023early} and low-resource community screening initiatives in India~\citep{dubey2024mobile}, but these datasets typically remain within research groups to protect participant privacy and retain competitive advantage. Similarly, crowdsourced video datasets used for validating online screening approaches~\citep{washington2019validity} have shown promise but face quality control challenges and ethical considerations around public data sharing.
\subsubsection{Comparative Analysis}
Table~\ref{tab:dataset_comparison} provides a systematic comparison of the key characteristics of major ASD behavioral datasets. Critical dimensions for comparison include dataset size (number of subjects and videos), age range of participants, diagnostic information available (at-risk vs. confirmed diagnoses), modalities captured (video, audio, eye-tracking), level of annotation granularity (video-level, event-level, or frame-level), temporal coverage (single time point vs. longitudinal), and data availability (public, restricted, or private).
This comparison reveals several consistent patterns across existing datasets. Most datasets are relatively small, with participant counts typically in the tens to low hundreds. Longitudinal designs tracking the same children over multiple developmental time points are rare, with most datasets capturing single assessments. Frame-level behavioral annotations, essential for training temporal models for behavior detection, are available in only a subset of datasets. Finally, the tension between data sharing for scientific progress and protection of participant privacy has resulted in many valuable datasets remaining restricted or unavailable to the broader research community~\citep{mukherjee2024digital}.
\subsubsection{Key Gaps and Limitations in Existing Datasets}
Synthesizing across existing resources reveals critical limitations that hinder the development of clinically deployable screening tools. Foremost among these is the issue of scale; most datasets contain fewer than 200 participants, offering insufficient statistical power for deep learning models that require large- scale datasets to generalize effectively~\citep{litman2025decomposition}. This scarcity is compounded by a lack of longitudinal data tracking developmental trajectories from infancy~\citep{Ozonoff2019} and a dearth of frame-level annotations, which are essential for providing the temporal supervision needed to train precise behavior detectors. Furthermore, many existing datasets lack ecological validity, relying on structured tasks or specialized equipment that do not reflect the naturalistic variability of real-world clinical settings~\citep{mukherjee2024digital}. These technical gaps are exacerbated by modality constraints that prevent true multimodal integration and stringent privacy barriers that keep valuable data (such as proprietary mobile health datasets) in silos, preventing broader community validation. In addition, fine-grained behavior detection in monocular social interaction videos remains challenging, particularly under non-ideal conditions such as non-frontal views, occlusions, and unknown camera parameters. Compared to appearance-based approaches \citep{marin2019laeo,chong2020detecting,tafasca2023childplay}, feature-based methods leveraging low-dimensional, behavior-specific representations offer a computationally efficient alternative for detecting subtle behavioral patterns.
The ASD-FEAT dataset, described in the following section, was designed with explicit attention to these challenges, offering a larger, longitudinally structured resource with comprehensive frame-level annotations and carefully-curated features in providing an efficient ML platform for identifying clinically relevant behaviors for ASD screening.

\section{ASD-FEAT Dataset}
\label{dataset}
ASD is characterized by impairment in social communication, interactive skills and repetitive behaviors. Basic behaviors such as gaze towards a conversation partner's face, smiling and vocalizing at key moments during social interactions have drawn much attention in autism research for their potential as early behavior markers of ASD. 
Group-level differences in the rate of occurrence of these behaviors can emerge as early as 9-12 months \citep{Ozonoff2019}. Specifically, by 9–12 months, typically developing (TD) infants show high rates of gaze to faces, which is stable over time, as well as increasing rates of social smiling and social vocalizations as they get older; in contrast, infants later diagnosed with ASD exhibit declines across all three behaviors from 6 to 36 months.

\subsection{Behavioral Markers of Social Interaction}
The ASD-FEAT dataset includes a carefully selected set of behavior annotations capturing key social interaction markers, defined and labeled by experienced researchers. These markers, including Look Face, Look Object, Smile, Vocalization, Social Smile, and Social Vocalization, are crucial for identifying key social behaviors in children and serve as the foundation for downstream ML analysis for early ASD risk prediction. Table~\ref{tab:behavioral_coding} summarizes the operational definitions applied during manual coding to ensure consistency and reliability.

\begin{table*}[t!]
\footnotesize
    \centering
    \caption{Behavioral Coding Definitions}
    \label{tab:behavioral_coding}

\begin{tabular}{ll}
\hline
\textbf{Behavior}            & \textbf{Operational definition applied during manual coding}                                                \\
\hline
Look Face           & \begin{tabular}[c]{@{}l@{}}(a) The child's face is on-screen and directly facing the partner's face, or \\ (b) a portion of the partner's body is visible to contextualize facial orientation, or \\ (c) a gaze shift coincides with an off-screen partner's voice onset.\end{tabular}                                                  \\
\hline
Look Object         & \begin{tabular}[c]{@{}l@{}}The child's gaze is directed towards objects that: \\(a) are part of the current experimental task, or \\(b) were previously utilized in the task, or \\(c) were presented by the examiner or parent during the session, or \\(d) serve as potential reinforcers (e.g., toy cars, comfort objects).\end{tabular} \\
\hline
Smile               & \begin{tabular}[c]{@{}l@{}}(a) Observable upturning of at least one mouth corner, or visible eye muscle engagement, or \\(b) audible expressions of joy such as chuckling or explosive vocal sounds. \\The manual coding distinguishes genuine positive affect from incidental \\facial movements.\end{tabular}                            \\
\hline
Vocalization        & \begin{tabular}[c]{@{}l@{}}(a) non-verbal sounds (open vowels, consonant-vowel combinations), or \\(b) single word or word approximation articulations, or \\(c) multi-word phrases, or \\(d) communicative air gasps signaling excitement. Non-communicative \\physiological sounds are excluded.\end{tabular}                      \\
\hline
Social Smile        & Any smile event co-occurring or occurring within a two seconds of a look face event                                                                                              \\
\hline
Social Vocalization& Any vocalization event co-occurring or occurring within two seconds of look face  \\
\hline
\end{tabular}

\end{table*}

\subsection{Dataset Description}

\subsubsection{Video Dataset Collection}
\label{feature_description}
ASD-FEAT consists of meta-data and derived feature data obtained from video recordings of table-based adult-child play interactions in a clinical setting. The source videos were recorded between 2009 and 2019 as part of an IRB-approved longitudinal infant sibling study, in which 318 children were  assessed at 6, 9, 12, 18, 24, and 36 months of age (refer to Table~\ref{tab:dataset_demographic_spec} for demographic information). The cohort includes infant siblings of children with ASD as well as typically developing children. This sampling strategy resulted in a higher prevalence of ASD (17\%) compared to general population estimates ($\sim$3.2\%). The proportion of male participants varies across age groups from 47\% to 59\%, with an overall average of 54\%.

\begin{table}[]
\centering
\caption{Sample sizes for each assessment timepoint (6, 9, 12, 18, 24, 36 months). The dataset, comprising of 318 unique subjects, is longitudinal, featuring visits at different ages with up to three interactive play sessions recorded per visit for most subjects. In the age rows, all counts (including ASD/Non-ASD) are numbers of recorded play interactions and sum to Total N.}
\label{tab:dataset_demographic_spec}
\begin{tabularx}{\linewidth}{l *{8}{>{\centering\arraybackslash}X}}
\hline
Age (m) & Total N & \multicolumn{2}{c}{Interactive play with} & Male & Female & ASD & Non-ASD & \% ASD  \\
\cline{3-4}
        &         & \multicolumn{1}{c}{Examiner} & \multicolumn{1}{c}{Parent} & & & & &  \\
\hline
6       & 334     & 84       & 250    & 181  & 153    & 0   & 334     & 0\%     \\
9       & 140     & 0        & 140    & 82   & 58     & 0   & 140     & 0\%     \\
12      & 477     & 119      & 358    & 259  & 218    & 4   & 473     & 0.8\%  \\
18      & 312     & 112      & 200    & 148  & 164    & 38  & 274     & 12.2\% \\
24      & 293     & 98       & 195    & 157  & 136    & 41  & 252     & 14.0\% \\
36      & 330     & 115      & 215    & 173  & 157    & 57  & 273     & 17.3\% \\
Total   & 1886    & 528      & 1358   & —    & —      & 54  & 264     & 17.0\% \\
\hline
\end{tabularx}
\end{table}

\begin{figure}[t]
    \centering
    \begin{subfigure}[b]{0.48\textwidth}
        \centering
        \includegraphics[width=\textwidth]{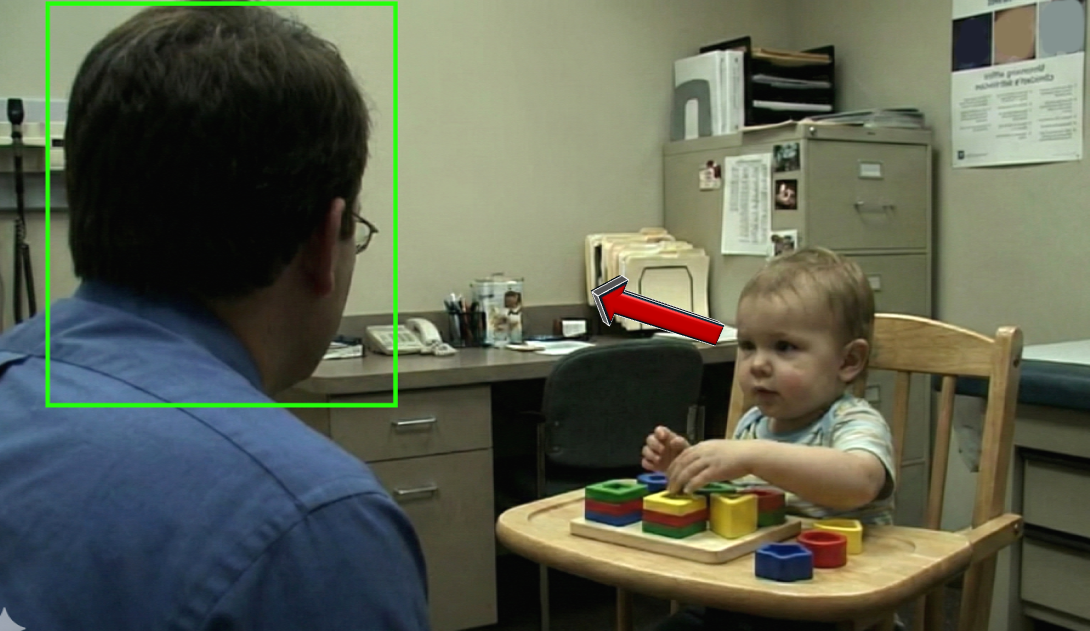}
        \caption{}
        \label{fig:head}
    \end{subfigure}
    \hfill 
    \begin{subfigure}[b]{0.45\textwidth}
        \centering
        \includegraphics[width=\textwidth]{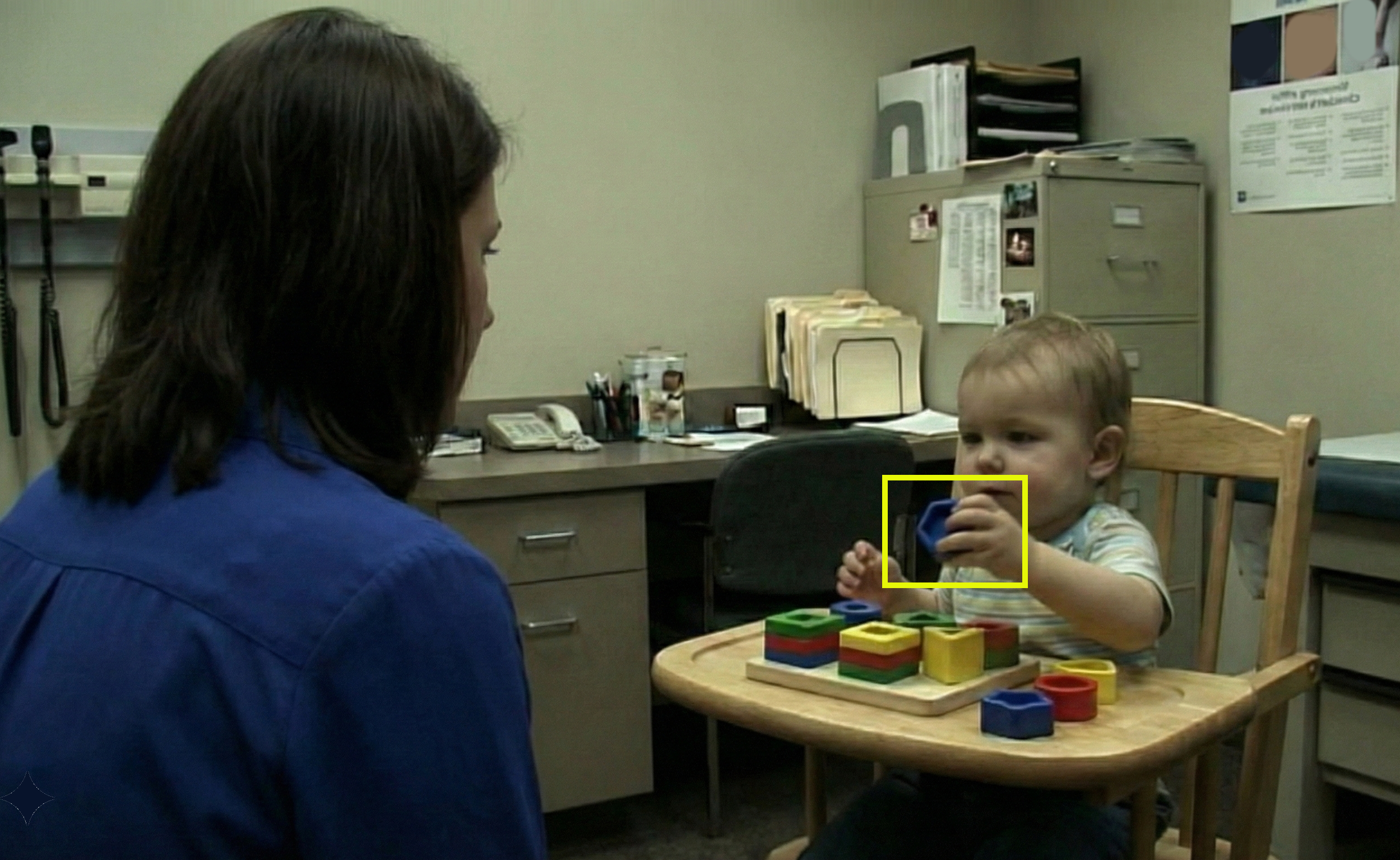}
        \caption{}
        \label{fig:toy}
    \end{subfigure}
    
    \caption{Visual representation of the interaction showing (a) the bounding box detection of the adult's head and vector representing child's gaze in 3D and (b) Isolation of the toy as a dynamic object using OpenCV background subtraction. 
    To protect subject privacy, the images in this figure are generated examples rather than frames from actual clinical videos. }
    \label{fig:main_figure}
\end{figure}

Each video sample is a recording of a 3-minute interactive play session with either a parent or an examiner as shown in Figure~\ref{fig:main_figure}. The experiment is designed to study the interplay of interaction between children and adult. During the task, the child participates in a series of activities and tasks intended to elicit social responses, visual attention and vocalization. A variety of toys and manipulatives are used to keep the child engaged. The adult may also introduce new toys to observe the child's response to different stimuli. Each session is recorded using two cameras: one capturing the child’s face and actions and the other recording the adult’s face, both in semi-profile view. The videos were manually annotated by trained coders, achieving at least 80\% inter-rater agreement on the onset and offset times and durations of all behavior events defined in Table~\ref{tab:behavioral_coding}. To ensure reliability, 15\% of the data underwent double coding by master coders, yielding high intra-class correlation coefficients (ICCs) above 0.90 for all behavior codes. In addition, the participants were tested by expert clinicians using a comprehensive diagnostic battery. Based on the results of the full evaluation, subjects were assigned to ASD or non-ASD outcome groups at each visit. Final classifications were determined at the 36-month visit, based on whether they met formal DSM-5 diagnostic criteria at that timepoint or not. 

All data collection and analysis adhered to ethical guidelines after obtaining informed consent from parents for the use of video data, with strict measures implemented to ensure participant privacy, including the use of de-identified study-local subject identifiers and the decision not to release any raw video or image data of child participants. Together, these design choices provide a longitudinal, multimodal dataset with clinically validated outcomes and high-quality annotations, supporting the development of robust ASD screening.

The following subsections describe the derived feature representations extracted from the source videos that constitute the released ASD-FEAT dataset. Table~\ref{tab:dataset_feat_spec} summarizes the field names and groupings for all derived features (gaze and pose, facial landmarks and action units, object-of-interest bounding boxes, and raw I3D and mel-spectrogram representations), and Table~\ref{tab:dataset_labels_spec} enumerates the accompanying frame-level ground-truth behavior labels and metadata fields.

\newcolumntype{L}{>{\raggedright\arraybackslash}X}

\begin{table}[ht]
\small
\centering
\caption{Comprehensive specification of features in the ASD-FEAT dataset.}
\label{tab:dataset_feat_spec}
\begin{tabularx}{\linewidth}{p{2.2cm} p{2.5cm} L}
\toprule
\textbf{Category} & \textbf{Sub-Type} & \textbf{Field Details \& Description} \\ \midrule
\multirow{2.5}{*}{Gaze \& Layout} & 3D Gaze & \texttt{gaze0-2}: 3D gaze direction vector from Gaze360. \\ \cmidrule{2-3}
 & Adult Head & \texttt{tl\_x, tl\_y, br\_x, br\_y}: Bounding box coordinates for the adult partner's head. \\ \midrule
\multirow{3.5}{*}{Facial} & Landmarks & \texttt{C\_x/y\_0...67}: 68-point 2D facial landmarks (jaw, nose, eyes, lips). \\ \cmidrule{2-3}
 & Eye Lmk. & \texttt{C\_eye\_lmk\_x/y}: 56-point 2D landmarks for the eye region. \\ \cmidrule{2-3}
 & Action Units & \texttt{C\_AUxx\_r/c}: 17 AU intensities and 18 AU presence flags (presence set includes \texttt{AU28\_c}, lip suck, with no corresponding intensity channel). \\ \midrule
Scene & Objects & \texttt{toytlx, toytly, toybrx, toybry}: Bounding box coordinates for toys/objects. \\ \midrule
\multirow{2.5}{*}{Raw Data} & Video & \texttt{I3D RGB/Flow}: Spatiotemporal embeddings from I3D model. \\ \cmidrule{2-3}
 & Audio & \texttt{Mel Spectrogram}: Mel-spectrogram representations (48kHz). \\ \bottomrule
\end{tabularx}
\end{table}

\begin{table}[ht]
\small
\centering
\caption{Ground truth labels available in the ASD-FEAT dataset.}
\label{tab:dataset_labels_spec}
\begin{tabularx}{\linewidth}{p{2.2cm} l L} 
\toprule
\textbf{Category} & \textbf{Sub-Type} & \textbf{Description (Field Names Included)} \\ \midrule
\multirow{4}{*}{Ground Truth} & Behavior Signals & \textbf{look\_face, look\_object, smile, vocal}: Frame-level binary labels for gaze target, social smiles, and vocalizations. \\ \cmidrule{2-3}
 & Frame Metadata & \textbf{frame, timestamp, valid, C\_confidence}: Frame indices, validity flags, and tracking confidence scores. \\ \cmidrule{2-3}
 & Session Metadata & \textbf{sub\_id, visit, partner}: De-identified study-local subject identifier; visit age in months (6, 9, 12, 18, 24, 36); interacting adult for the session (Examiner or Parent). \\ \cmidrule{2-3}
 & Diagnostic Outcome & \textbf{asd\_longitudinal, asd\_36}: Best-estimate clinical diagnosis (ASD or Not ASD) at each visit (\texttt{asd\_longitudinal}) and the final 36-month outcome (\texttt{asd\_36}). \\ \bottomrule
\end{tabularx}
\end{table}

\subsubsection{Face and Eye Landmarks Detection}

\begin{figure*}[t]
\centering\includegraphics[width=\textwidth, trim=0 0 0 0, clip]{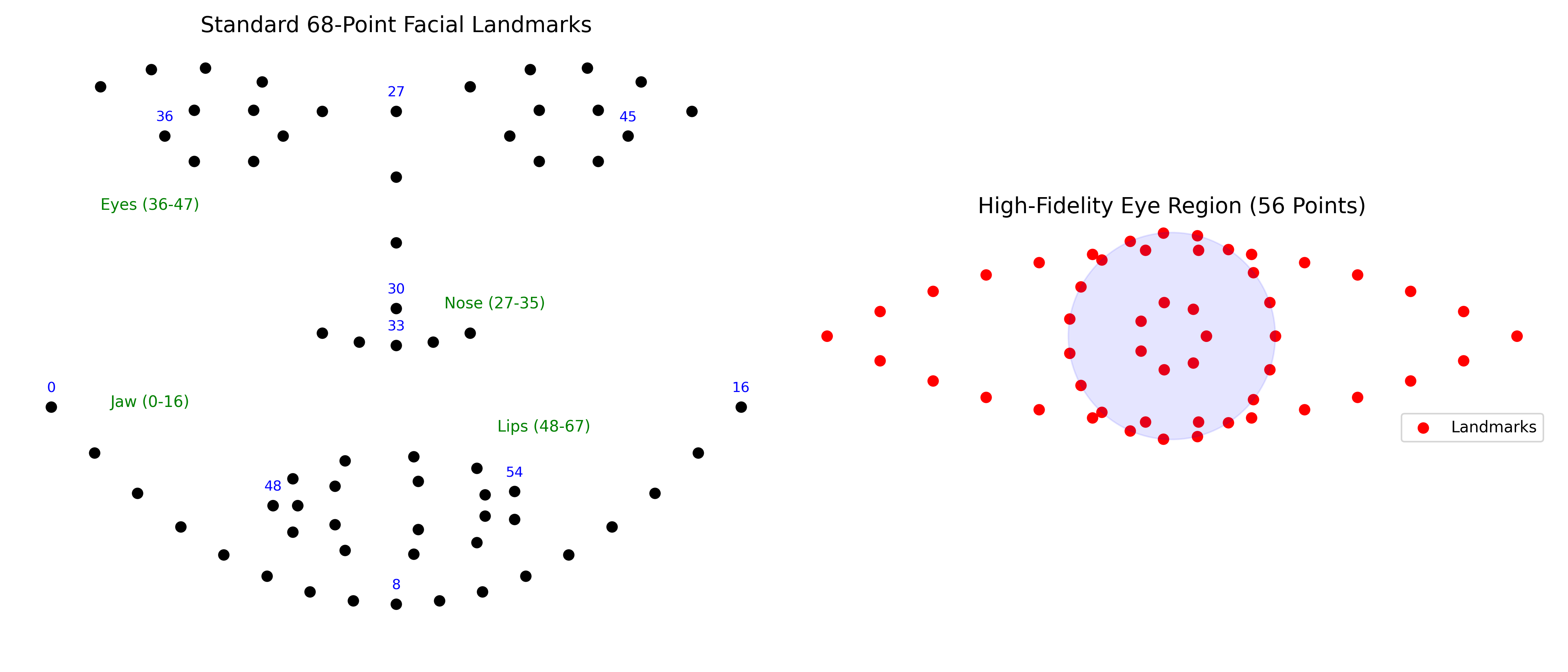}
\caption{(Left) The standard 68-point facial landmark configuration used for tracking the jawline, eyebrows, nose, and mouth. (Right) A detailed 56-point eye landmark model, including eyelid contours and iris boundaries, utilized for robust gaze estimation.}
\label{fig:landmarks}
\end{figure*}

For facial landmark detection from video, the Convolutional Experts Constrained Local Model (CE-CLM)\citep{zadeh2017convolutional} is applied using the OpenFace 2.0 package \citep{baltrusaitis2018openface}. CE-CLM uses a Multi-task Convolutional Neural Network (MTCNN) for face detection initialization, followed by a four layer CNN network for tracking faces of interest in subsequent frames. For eye landmark detection, OpenFace uses a Constrained Local Neural Field (CLNF) landmark detector \citep{baltrusaitis2013constrained}. 
 The child's face is represented by the 68 2D face landmarks $F = \{f_1, f_2, ... f_{68}\}$ and 56 2D eye landmarks $E = \{e_1, e_2, ... e_{56}\}$, each defined by (x,y) in image space (Figure~\ref{fig:landmarks}).

\begin{sloppypar}
While more recent landmark detection methods offer improved accuracy on modern benchmark datasets, we intentionally selected OpenFace for several principled reasons aligned with our dataset design goals. First, OpenFace provides good stability and robustness for the challenging recording conditions in our naturalistic clinical videos such as non-frontal head poses, variable distances from camera, and child movement, compared to newer methods that are often trained and optimized for more controlled scenarios such as driver monitoring (frontal, dual-camera setups) or laptop webcam applications (frontal, close-range, high-quality video). The videos exhibit the quality characteristics and compression artifacts reflective of comsumer-level production rather than the high-resolution, well-lit conditions assumed by contemporary deep learning approaches. Second, OpenFace's 68-point facial and 56-point eye landmark representations provide sufficient granularity for behavioral analysis while maintaining participant privacy through lower-dimensional representations. Modern methods employing hundreds or thousands of landmark points may increase re-identification risk and contradict our privacy-preserving dataset design philosophy. Finally, our primary objective is enabling reproducible research through release of stable, well-documented feature representations. OpenFace's maturity, extensive validation, open-source availability, and widespread adoption in the behavioral research community make it the practical choice for maximizing dataset utility and research accessibility. Future work can explore whether features from newer landmark detectors improve downstream behavioral classification, but such exploration depends on the availability of foundational datasets like ASD-FEAT.
\end{sloppypar}

\subsubsection{Gaze Direction Estimation}
Computer vision--assisted methods for objective gaze measurement in ASD risk prediction have been attempted in varying configurations. Automatic detection of behaviors like mutual gaze or gaze towards object of interest is complicated by challenging conditions such as non-frontal views, partial visibility of the face, and the lack of depth information and camera calibration. Effective gaze estimation typically integrates multiple sources of information, including facial and eye landmarks, head pose, and the subject's 3D gaze vector, to estimate the direction of attention toward people and objects in the scene.

Accurately estimating 3D eye gaze is important for reliably identifying subtle social behaviors such as gaze direction toward faces or objects. To address this, we employed Gaze360, a state-of-the-art 3D gaze estimation model designed for use in unconstrained, real-world environments \citep{kellnhofer2019gaze360}. In our experiments, substituting 2D gaze features with Gaze360-derived 3D gaze vectors (Figure~\ref{fig:head}) led to a significant improvement in detecting “look” behaviors, increasing overall classification accuracy by 10\%.

\subsubsection{Head Position Detection}
The videos are recorded from an over-the-shoulder perspective, with only the back of the adult participant's head visible. To detect a mutual gaze event, the head position of the interacting partner is required, in addition to the position of the child's head and gaze direction. We use a Single Shot Multibox Detector (SSD) based head detector as described in \citep{liu2016ssd}, and trained on the Hollywood Movies dataset \citep{klaser2010human} and the EgoHand dataset \citep{Bambach_2015_ICCV}. For implementation, we adopt the Keras-based SSD framework developed by Pierluigi Ferrari \citep{ssd_keras}, specifically from the project \citep{ssd_head_keras}. The head position is represented as $H = \{h_1, h_2, h_3, h_4\}$ which represent the 2D coordinates of the top left and bottom right corners of the detected head region (Figure~\ref{fig:head}).

\subsubsection{Object-of-Interest (toy) Position Detection}

To provide spatial context for estimating a child's visual focus, we incorporate the spatial location of target objects (toys) within the video frames. Rather than employing a dedicated object detection model for toys as a preliminary step, we utilized background subtraction to identify regions of motion within the scene. To exclude motion from upper body regions, we first masked out regions corresponding to the child's face and body, as well as any adults present in the frame. This allowed the background subtraction to specifically isolate motion from hands and dynamic objects (e.g., toys). While this inherently captures all other dynamic elements, it concurrently isolates instances of toy motion when present. The detected motion regions are represented to the model as bounding boxes represented as $T = \{t_1, t_2, t_3, t_4\}$ which represent the 2D coordinates of top left and bottom right corners of the detection region respectively, providing a spatial cue for potential object locations alongside the child's facial landmarks and their 3D gaze direction (Figure~\ref{fig:toy}).

\subsubsection{Facial Action Units (AUs) Estimation}
Facial expression dynamics, specifically reduced frequency of social smiles are well-established behavioral markers for ASD. There is an association between reduced social smiling and later autism outcomes \citep{ozonoff2010prospective}. Quantitative coding studies have demonstrated that children with ASD exhibit statistically significant differences in spontaneous smile generation, social referencing, and interpersonal affect modulation compared to neurotypical developmental trajectories. Objective measures further validate that these specific gaze and smile dynamics reflect core ASD-related characteristics \citep{ahn2024objective}.

To capture nuanced facial expressions and augment the smile detection model, our approach leverages estimates for a subset of Facial Action Units (AUs) provided by the OpenFace software,  which are based on the Facial Action Coding System (FACS). FACS offers a standardized, objective taxonomy for human facial movements, deconstructing expressions into specific AUs that correspond to individual muscle movements. OpenFace specifically detects AUs 1, 2, 4, 5, 6, 7, 9, 10, 12, 14, 15, 17, 20, 23, 25, 26, 28, and 45, providing both their presence (binary occurrence) and intensity (on a 5-point scale). The system combines appearance-based features (e.g. Histogram of Oriented Gradients (HOG)) with geometry-based features derived from facial shape parameters and landmark locations. The facial action units features are represented as $Z_i = \{z_1, z_2, ... z_{35}\}$ where each $\{z_i\}$ encodes the presence (0/1) and intensity of a specific action unit.

\subsubsection{I3D RGB and flow Features}
I3D (Inflated 3D ConvNet) \citep{carreira2017quo} is a deep learning architecture that processes both spatial and temporal information through two separate streams networks for RGB and optical flow features. We extract RBG and flow features from the ASD videos by running inference on the I3D model pre-trained on the Kinetics dataset. We use non-overlapping windows of 16 consecutive frames as inputs to the I3D model, resulting in 1024 features for both the RGB and flow streams.

\subsubsection{Mel Spectrogram}

While the visual features are indispensable in detecting facial expressions and eye gaze, audio features provide a more direct measurement of vocalization behaviors. To detect vocalization, audio stream provides a more direct cue. We convert the audio into a joint time-frequency representation in the form of mel spectrogram using the librosa library~\citep{mcfee2015librosa}. The input audio signal, sampled at 48 kHz is first segmented into short frames of 1024 samples, with a hop length of 240. Short-Time Fourier Transform (STFT) is applied to each frame to convert it from the time domain to the frequency domain, revealing the strength of different frequencies over time with output mapping on 2048-point Short Time Fourier Transform (STFT). Then, a Mel filter bank, which approximates human perception of pitch by spacing filters logarithmically, is applied to the linear frequency spectrogram to map the frequencies to the mel scale with $M=112$ coefficients. The resulting Mel spectrogram, denoted as $S$, is a matrix where the rows correspond to the Mel filter bands and the columns correspond to the temporal frames. The number of columns (audio frames) is determined by the total number of audio samples, the hop length, and the frame duration. For a 3-minute audio signal ($T=180$ seconds), the output Mel spectrogram $S$ has dimensions $(M, N_{frames}) = (112, 36001)$. These audio features are included in ASD-FEAT, allowing researchers to analyze vocal behaviors without requiring access to raw audio recordings.

The feature representations described above provide a multi-faceted characterization of social behavior spanning geometric (landmarks, gaze, position), affective (action units), auditory (mel spectrograms), and motion (I3D) modalities. These complementary features enable our system to capture the subtle behavioral differences that distinguish children with ASD from typically developing peers. We next describe the machine learning framework that operationalizes these features for automated behavior detection and ASD risk assessment.

\subsection{Licensing, Access, and Ethics}
\label{sec:licensing}

\subsubsection{Ethics and Source Data Provenance}
The source video recordings underlying ASD-FEAT were collected as part of an IRB-approved longitudinal infant sibling study at a U.S. clinical research institute. Written informed consent was obtained from a parent or legal guardian of every participating child for use of the recordings in research, including secondary analysis.

\subsubsection{Released Features and Disclosure-Risk Mitigation}
We will release only derived machine-learning features (facial and eye landmarks, head and gaze pose estimates, action units, I3D embeddings, and mel spectrograms), each keyed by a de-identified study-local subject identifier, together with frame-level behavioral labels, de-identified metadata (child age at visit and child sex), the diagnostic outcome (ASD or Not ASD) at each visit, and the final 36-month diagnostic outcome (ASD or Not ASD). These derived features exclude the original video pixels and audio waveforms, and the raw video and audio recordings will not be redistributed under any circumstances. Several technical properties of the released features substantially reduce disclosure risk beyond nominal de-identification, although some residual risk may remain. The face and eye landmark data comprise only sparse geometric landmark coordinates together with derived head-position and gaze estimates; these encode relative geometry and contain no pixel, texture, or appearance information, and their limited resolution and point count substantially reduce the likelihood that a recognizable image of any participant could be reconstructed. No GUIDs, dates of birth, or exact ages are released; age is reported only as the study's assessment ages (6, 9, 12, 18, 24, and 36 months). The audio features are reduced-resolution mel-spectrogram summaries whose limited spectral and temporal detail substantially reduces the risk of speaker re-identification. The I3D features are high-level, abstract embeddings from an action-recognition model that summarize generalized movement patterns rather than visual content and are not readily invertible to recover identifiable video.

\subsubsection{Dataset License}
The dataset is released under an open-access license, with no data use agreement or controlled-access gating. The license terms prohibit any attempt to re-identify participants, whether by linkage to external sources or by adversarial reconstruction from features; restrict usage to non-commercial research consistent with the original participant consent; require citation of this work; and limit redistribution. Redistribution of the features in derived form is permitted under the same terms; redistribution alongside reconstructed audio or video is prohibited.

\subsubsection{Upstream Tool and Model Licenses}
ASD-FEAT features are produced using third-party tools whose licenses propagate to downstream users. Table~\ref{tab:upstream_licenses} summarizes these dependencies. Users intending commercial deployment should review each upstream license individually, as several restrict use to non-commercial research.

\begin{table}[h]
\small
\centering
\caption{Upstream tools and pretrained models used to construct ASD-FEAT features, with licensing terms relevant to downstream redistribution and use.}
\label{tab:upstream_licenses}
\begin{tabularx}{\linewidth}{p{2.6cm} p{2.4cm} L}
\toprule
\textbf{Tool / Model} & \textbf{License} & \textbf{Scope} \\ \midrule
OpenFace 2.0 \citep{baltrusaitis2018openface} & Custom academic, non-commercial & Facial landmarks, eye landmarks, head pose, action units \\ \midrule
Gaze360 \citep{kellnhofer2019gaze360} & CC BY-NC 4.0 & 3D gaze direction estimation \\ \midrule
I3D pretrained on Kinetics \citep{carreira2017quo} & Apache 2.0 (code); Kinetics labels CC BY 4.0 & RGB and optical-flow spatiotemporal embeddings \\ \midrule
librosa \citep{mcfee2015librosa} & ISC & Mel-spectrogram extraction from session audio \\ \midrule
ResNet-50 (vocalization classifier backbone) & BSD 3-Clause (torchvision) & Mel-spectrogram--based vocalization detection \\ \midrule
DBSCAN (scikit-learn) \citep{ester1996density} & BSD 3-Clause & Event clustering on frame-level probability streams \\ \bottomrule
\end{tabularx}
\end{table}

\subsubsection{Access and Hosting}
To further protect participant privacy, ASD-FEAT will be deposited in Zenodo, a publicly funded, FAIR-compliant (Findable, Accessible, Interoperable, Reusable) repository, under the persistent DOI \url{https://doi.org/10.5281/zenodo.22261227}, with access managed through the repository's approved procedures rather than by the study team. Access details will be made available at the project page (an anonymized data sample is provided to reviewers as supplementary material). The release will include the feature files, frame-level annotation files, a datasheet for the dataset (Appendix~\ref{app:datasheet}), and reference scripts for reproducing the experiments reported in Sections~\ref{sec:framework} and~\ref{sec:experiments}.

\subsubsection{Maintenance and Versioning}
The dataset will be versioned with semantic version tags; corrections to annotations or feature files will be released as patch versions, while extensions to additional age timepoints or feature modalities will be released as minor versions. Each version is assigned its own DOI to ensure citation stability.

\section{ASD Risk Prediction Framework}
\label{sec:framework}

\subsection{Overview}

We propose the ASD risk prediction framework shown in Figure~\ref{fig:pipeline}. Following pre-processing, the machine learning stage detects fine-grained behaviors from adult-child interactive play videos using transformer-based classifiers. These behaviors -- gaze to faces, gaze to objects, social smiles, and social vocalizations, are known to correlate with ASD risks~\citep{ozonoff2010prospective}. Visual features such as facial landmarks, head pose, and 3D gaze serve as inputs to the model, enabling behavior identification without requiring highly controlled environments or camera calibration.

Detected behaviors are represented as time series, encoding their frequency, duration, and co-occurrence. In the final stage, features derived from these time series, including rates and durations of all the behaviors of interest listed in Table~\ref{tab:behavioral_coding} are used as inputs to a classifier for ASD risk prediction.

Our approach treats behavior detection and risk prediction as a two-stage cascade. The first stage (Section~\ref{sec:Behavior_from_video}) produces interpretable behavioral metrics---rates and durations of all the behaviors of interest listed in Table~\ref{tab:behavioral_coding}---that summarize the child's social engagement profile during the 3-minute interaction. The second stage (Section~\ref{sec:risk_prediction}) maps these behavioral profiles to ASD risk using supervised classification. This decomposition offers several advantages: (1) it enables comparison between machine-derived and manually coded behaviors, thereby establishing the fidelity of automated detection, and (2) it provides clinically interpretable intermediate representations, and allows the risk prediction model to be retrained or recalibrated without reprocessing videos.

\subsection{Stage 1: Behavior Detection from Video}
\label{sec:Behavior_from_video}

\subsubsection{Automatic Behavior Coding from Visual Features}
We employ a self-attention based transformer architecture that operates on a temporal sequence of multimodal features. These features are extracted using standard facial behavior analysis tools, thus reducing dependence on camera setup and calibration in the recording environment. Additionally, the features offer visual and semantic context, enhancing the interpretability of the classification model's performance~\citep{murdoch2019interpretable}.

\subsubsection{Transformer Architecture}

Transformer is a self-attention--based architecture that has become dominant in natural language processing \citep{vaswani2017attention}. It has been successfully adapted for computer vision tasks including object detection \citep{carion2020end} and mutual gaze detection \citep{guo2022mgtr}.

The transformer architecture offers significant advantages for detecting behaviors in social interaction videos. Unlike approaches that rely on CNNs with fixed window sizes, the self-attention mechanism enables the model to dynamically learn which features across the sequence are most relevant. This allows the model to capture long-range dependencies—for instance, a subtle eye movement in one frame may signal a look-to-face event occurring several frames later.

\begin{figure*}[t!]
\centering\includegraphics[width=\textwidth, trim=0 0 0 0, clip]{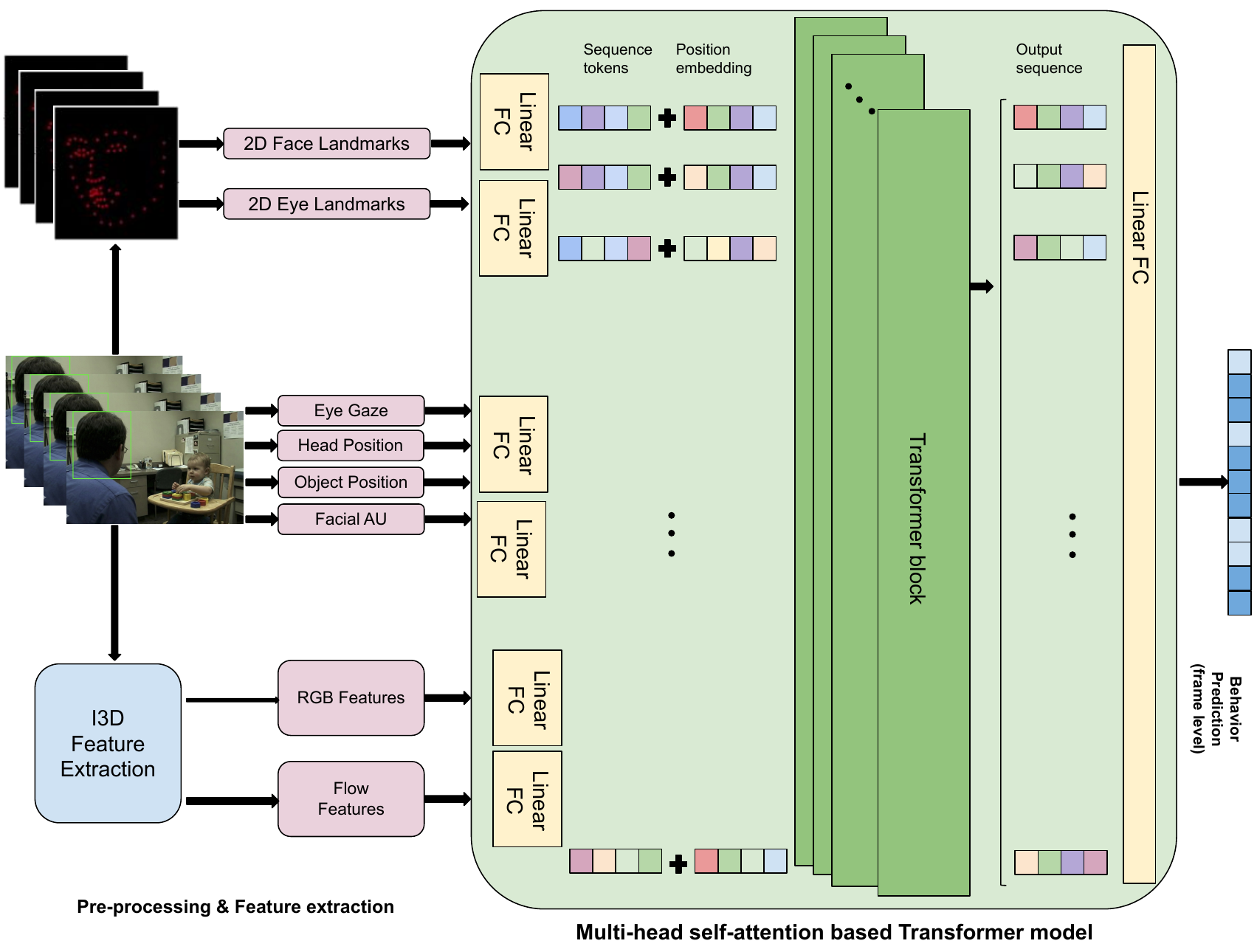}
\caption{Input features (face and eye landmarks, facial AUs, head and object position, eye gaze) are encoded through MLP and concatenated together to form input to the transformer model. The transformer model output passed through an MLP resulting in frame level target behavior prediction. To protect subject privacy, the images in this figure are generated examples rather than frames from actual clinical videos.}
\label{fig:transformer_model}
\end{figure*}

\paragraph{Input Feature Construction} Our proposed behavior detection model (Figure~\ref{fig:transformer_model}) is an encoder-only Transformer architecture designed for fine-grained frame-level analysis. The model processes a sequence of $N$ frames, where the input for each frame is not raw pixel data, but a structured composition of derived behavioral features. To construct the input token for a given frame t, we utilize a diverse set of hand-crafted features (as detailed in Section \ref{feature_description}). 

To handle the varying dimensions of these heterogeneous features, each feature type is first passed through a dedicated linear layer (MLP) to project it into a lower-dimensional representation: head pose ($11 \rightarrow 6$), facial landmarks ($136 \rightarrow 30$), eye landmarks ($112 \rightarrow 30$), action units ($35 \rightarrow 10$), I3D RGB ($64 \rightarrow 16$), and I3D optical flow ($64 \rightarrow 16$). These projected features are then concatenated to form the final input vector $v_t \in \mathbb{R}^k$, where $k = 108$.

Given an input video sequence with corresponding feature vectors $X = \{\mathbf{x_1}, \mathbf{x_2}, \ldots, \mathbf{x_N}\}$, the projection and concatenation process yields a sequence of embedding vectors $V = \{\mathbf{v_1}, \mathbf{v_2}, \ldots, \mathbf{v_N}\}$, where each $\mathbf{v_t} \in \mathbb{R}^k$ and $k$ is the embedding dimension. To preserve temporal order within the self-attention mechanism, we incorporate sequence indices through learned positional embeddings. These positional embeddings $P = \{\mathbf{p_1}, \mathbf{p_2}, \ldots, \mathbf{p_N}\}$ have the same dimension as the feature embeddings and are combined through element-wise addition, such that the final input to the Transformer encoder is $z_t = v_t + p_t$.

We use an 8-layer, 8-head Transformer to provide sufficient capacity for modeling temporal dependencies while avoiding over-parameterization given the dataset scale. The output of the transformer is the sequence of vectors $Y = \{\mathbf{y_1}, \mathbf{y_2}, ...\mathbf{y_t}\}$ where $y_i$ is the probability score of frame $i$ containing the target behavior, making this a binary classification task. Each output vector is a computed through attention-weighted combinations of all input embeddings.

\paragraph{Self Attention Mechanism}
The self-attention mechanism computes a weighted average of features with the weight proportional to a similarity score between pairs of input features. For a given input sequence, we compute Query (Q), Key (K), and Value (V) matrices through learnable linear projections. The attention scores are calculated as:
\begin{equation}
\mbox{Attention}(Q, K, V) = \mbox{softmax}\left(\frac{QK^T}{\sqrt{d_k}}\right)V
\label{eq:attention}
\end{equation}
where $d_k$ is the scaling factor based on the dimension of the key vectors. This operation allows the model to dynamically weight the importance of every frame j when constructing the representation for frame i, effectively capturing long-range temporal dependencies such as the onset of a smile or the duration of a gaze shift.

\paragraph{Training Loss}
The model outputs the prediction probability score representing the likelihood of the target behavior being present in frame $i$. This frames the problem as a binary classification task. To handle class imbalance inherent in behavioral data, we employ a class-weighted binary cross-entropy loss, $L_{cls}$. The model is trained independently for each target behavior using frame-level labels derived from expert annotations.

We train the model using the Adam optimizer with $\beta_1$ = 0.9 and $\beta_2$  = 0.999, a learning rate of 0.001, and weight decay of 0.001. Training proceeds for 50 epochs with a batch size of 16, using a temporal block size of 16 frames. These hyperparameters were selected based on standard practices in transformer-based sequence modeling and validated empirically to provide stable convergence without overfitting given the dataset size. The relatively short temporal window (N=16) balances capturing local behavioral dynamics with computational efficiency, while deeper temporal dependencies are captured through the clustering process described in the next section.

\subsubsection{Event Generation via Clustering} 

While the transformer model outputs frame-level probabilities $P(t) \in [0,1]$ for each behavior, clinical risk assessment requires higher-level event metrics. We convert continuous probability streams into discrete behavioral events using a Density-Based Spatial Clustering of Applications with Noise (DBSCAN)~\citep{ester1996density} based clustering approach that robustly handles brief dropouts and tracking noise. Frames with probability scores above threshold $\tau=0.5$ are marked as candidate event frames. To bridge temporal gaps caused by decision boundary instability while maintaining sensitivity to brief social signals, we apply DBSCAN to the temporal indices of candidate frames with tuned parameters $\epsilon=35$ and min\_samples=8 (Table~\ref{tab:dbscan_table}). At the 30\,Hz frame rate of the recordings, an $\epsilon=35$ allows for the merging of detections separated by up to $\sim$1.17 seconds, tolerating brief tracking dropouts and momentary occlusions during sustained behaviors such as a continuous look or an extended vocalization. We chose this window empirically to favor recovering complete behavioral episodes over fine-grained event splitting, since the downstream Stage 2 features (event count, duration, rate, and proportion) are more stable when behavioral episodes are kept intact. The min\_samples=8 constraint requires at least eight candidate frames ($\sim$0.27 seconds) per event, filtering out spurious, single-frame activations that do not constitute meaningful social behaviors. From the validated events, we compute four aggregate statistics per behavior: (1) Event Count (total number of event occurrences), (2) Total Duration (summed duration of all events), (3) Rate (event count divided by video length), and (4) Proportion (total event duration divided by video length).

\begin{table}[]
\footnotesize
\centering
\caption{Parameters for DBSCAN clustering}
\label{tab:dbscan_table}
\begin{tabular}{|l|l|l|}
\hline
Parameter             & Value   & Description                                                                                                                           \\ \hline
eps ($\epsilon$)      & \begin{tabular}[c]{@{}l@{}}35 frames\\ ($\approx$1.17 sec)\end{tabular} & \begin{tabular}[c]{@{}l@{}}Maximum temporal distance between candidate\\ frames to be considered neighbors; equivalently,\\ the largest gap bridged when merging detections\\ into a single event\end{tabular} \\ \hline
min\_samples          & \begin{tabular}[c]{@{}l@{}}8 frames\\ ($\approx$0.27 sec)\end{tabular}  & \begin{tabular}[c]{@{}l@{}}Minimum number of candidate frames to form\\ an event cluster; equivalently, the shortest\\ retained event duration\end{tabular}                                  \\ \hline
\end{tabular}
\end{table}

\subsection{Stage 2: ASD Risk Prediction from Behavioral Features}
\label{sec:risk_prediction}

Given the event-based behavioral statistics produced by Stage 1, the second stage maps each child's behavioral profile to an ASD risk prediction using supervised classification. We describe the construction of the feature vector, the candidate classifiers, our approach to class imbalance, and the evaluation protocol.

\subsubsection{Behavioral Feature Construction}
Each subject's 3-minute interaction is summarized by a fixed-length feature vector derived from the time-aligned behavioral events. For each of the six behaviors listed in Table~\ref{tab:behavioral_coding}, we compute the four aggregate statistics defined in Section~\ref{sec:Behavior_from_video} (event count, total duration, rate, and proportion). Concatenated with the subject's gender and age at visit, this yields a 26-dimensional feature vector per subject. The outcome variable is the best-estimate clinical ASD diagnosis, defined by a subject obtaining an ADOS score above the ASD cutoff and fulfillment of the DSM-5 criteria for ASD at the time of the visit.

\subsubsection{Classifiers}
We evaluate three classifiers for binary ASD risk prediction: Logistic Regression (LR), Random Forest (RF), and a fully connected Multilayer Perceptron (MLP). These span a range of inductive biases---linear decision boundaries (LR), axis-aligned recursive partitioning (RF), and continuous non-linear manifolds (MLP)---allowing us to assess which best suits the structure of the behavioral feature space.

\subsubsection{Class Imbalance Handling}
The dataset exhibits substantial class imbalance (93:7 Non-ASD to ASD). To establish a balanced training set, we combine the Synthetic Minority Over-sampling Technique (SMOTE) with Tomek Links under-sampling~\citep{batista2004study}. SMOTE synthesizes new minority-class samples by interpolating between nearest neighbors, while Tomek Links removes overlapping majority-class samples at the decision boundary. The choice of combining both techniques rather than applying either in isolation is supported by an similar ablation in literature~\citep{wu2021machine}, which reported that SMOTE alone improves sensitivity at a substantial cost to specificity (0.83 / 0.70), Tomek Links alone fails to meaningfully shift the decision boundary (0.08 / 1.00), and only the combined scheme achieves a favorable sensitivity--specificity trade-off (0.92 / 0.71).

\subsubsection{Evaluation Protocol}
\label{sec:eval_protocol}
We evaluate using stratified 10-fold cross-validation, stratifying folds over subject, gender, age, and ASD diagnosis. Folds are formed at the subject level, so all sessions from a given child fall in a single fold. A child contributes several sessions, across repeat visits and different interacting partners, and distributing these across folds would let the classifier identify individual children. No subject appears in both the training and test folds of the Stage 2 classifier, so the reported performance is free of subject leakage. Ground-truth and ML-estimated behaviors are evaluated on the same partition: 10\% of subjects train the Stage 1 behavior detectors, and the remaining 90\% are cross-validated for Stage 2 risk prediction. That partition is subject-disjoint, so no child contributes to both stages, and fixing it across labeling sources makes the ground-truth and ML results comparable. For cases where no useful features could be extracted from video, the ML splits consist of slightly fewer sessions than ground truth. We report sensitivity (proportion of true positive cases correctly identified), specificity (proportion of true negative cases correctly identified), positive predictive value (PPV; proportion of positive predictions that are true positives), AUROC, and accuracy, along with 95\% confidence intervals derived from the cross-validation folds.

These five metrics characterize the classifier on the research cohort, but two properties of that cohort limit their direct clinical interpretability. First, the cohort is prevalence-enriched ($\sim$7.5\% ASD in the full Stage~2 cohort and 14.7\% in the matched dual-partner subset) relative to the $\sim$3.2\% U.S.\ general-population rate, so the cohort-level PPV is systematically optimistic for a population-deployed screener. Second, under the severe class imbalance characteristic of ASD screening, raw accuracy is dominated by the majority class, and sensitivity and specificity must be combined to summarize diagnostic utility. To address both, Section~\ref{subsec:pop_scale} reports three additional metrics computed from the same cross-validation folds: PPV recalibrated to 3.2\% prevalence via Bayes' rule (PPV@3.2\%), which restores realism to the operating point at population scale; the positive and negative likelihood ratios $\mathrm{LR}^{+}=\mathrm{Sens}/(1-\mathrm{Spec})$ and $\mathrm{LR}^{-}=(1-\mathrm{Sens})/\mathrm{Spec}$, which are prevalence-independent functions of sensitivity and specificity that quantify how strongly each test result shifts post-test odds; and the Matthews correlation coefficient (MCC), which integrates all four confusion-matrix cells and avoids the accuracy bias that arises under severe class imbalance. The PPV recalibration mirrors the adjusted-PPV reporting practice for ASD digital-phenotyping screeners~\citep{perochon2023early}; $\mathrm{LR}^{+}$/$\mathrm{LR}^{-}$ and MCC are established diagnostic-evaluation indices from clinical epidemiology and the ML-on-imbalanced-data literature, respectively~\citep{jaeschke1994users, chicco2020advantages}. We keep the standard classifier metrics in the per-experiment tables (Tables~\ref{tab:ASD_classifier_ML}, \ref{tab:auroc_asymmetry}, \ref{tab:asymmetry_classifier}) for direct comparison across configurations, and consolidate the calibrated and imbalance-robust metrics in Section~\ref{subsec:pop_scale} for a single, coherent population-scale discussion.

\section{Experiments}
\label{sec:experiments}

We evaluate our proposed framework through a series of experiments focused on two key components: (1) fine-grained behavior classification from video using a transformer-based model, and (2) ASD risk prediction based on both ground-truth behavioral features (from expert manual coding) and machine-predicted behavioral features (from the transformer models) to assess the feasibility of fully automated screening. Together these experiments evaluate both the accuracy of automated behavior detection and its effectiveness in supporting early ASD screening.

\subsection{Behavior Detection Model Performance}
\label{subsec:behavior_model_perf}
\begin{table*}[]
\footnotesize
\centering
\small
\caption{Performance of our transformer-based model on Look Face, Look Object, Smile and Vocalization classification tasks. Model performance is evaluated incrementally as the following feature modalities are fused: Face \& Eye Landmarks, Facial Action Units, 3D Gaze, Head Position, Object Position, I3D (RGB and Flow), and Mel Spectrogram. For comparison, vocalization classification was performed using a ResNet50-based model.}
\label{tab:Behavior_model_performance}
\begin{tabular}{l|cccccll|ccc}
\hline
\begin{tabular}[c]{@{}c@{}}\textbf{Beha-}\\\textbf{vior} \end{tabular}
    & \multicolumn{7}{c|}{\textbf{Features}}                                      & \multicolumn{3}{c}{\textbf{Metrics}}            \\ \hline
 &
  \multicolumn{1}{c}{\begin{tabular}[c]{@{}l@{}}Face \& \\eye \\ lmks\end{tabular}} &
  \multicolumn{1}{c}{\begin{tabular}[c]{@{}l@{}}Facial \\ Action\\ Units\end{tabular}} &
  \multicolumn{1}{c}{\begin{tabular}[c]{@{}l@{}}Head \\ Position\end{tabular}} &
  \multicolumn{1}{c}{\begin{tabular}[c]{@{}l@{}}Object \\ Position\end{tabular}} &
  \multicolumn{1}{c}{\begin{tabular}[c]{@{}l@{}}3D \\ Gaze\end{tabular}} &
  \begin{tabular}[c]{@{}c@{}}I3D\\ (RGB,  \\Flow)\end{tabular} &
  \begin{tabular}[c]{@{}c@{}}Mel \\ Spec.\end{tabular} &
  \multicolumn{1}{c}{\begin{tabular}[c]{@{}l@{}}PPV\\ (\%)\end{tabular}} &
  \multicolumn{1}{c}{\begin{tabular}[c]{@{}l@{}}Recall\\ (\%)\end{tabular}} &
  \multicolumn{1}{l}{\begin{tabular}[c]{@{}l@{}}AUC\\ ROC\end{tabular}} \\ \hline
  
    & $\checkmark$ & $\checkmark$ &  $\checkmark$ &  &   &                       &                        & 36.2& 50.6& 0.67\\
\begin{tabular}[c]{@{}l@{}}Look\\ Face\end{tabular}             & $\checkmark$ & $\checkmark$ & $\checkmark$  & $\checkmark$ &  &                       &                        & 37.9& 57.2& 0.68\\
             & $\checkmark$ & $\checkmark$ & $\checkmark$ & $\checkmark$ & $\checkmark$ &                       &                        & 49.7& 57.1& 0.72\\
             & $\checkmark$ & $\checkmark$ & $\checkmark$ & $\checkmark$ & $\checkmark$ & \multicolumn{1}{c}{$\checkmark$} &                        & 48.4& \textbf{67.3}& \multicolumn{1}{c}{\textbf{0.76}} \\ \hline
\begin{tabular}[c]{@{}l@{}}Look\\ Object\end{tabular}     & $\checkmark$ & $\checkmark$ & $\checkmark$  &  &   &                       &                        & 84.8& 59.2& 0.64\\
             & $\checkmark$ & $\checkmark$ & $\checkmark$ & $\checkmark$ & $\checkmark$ & \multicolumn{1}{c}{$\checkmark$} &                        & 86.3& \textbf{86.0}& \textbf{0.71}\\ \hline
\begin{tabular}[c]{@{}c@{}}Smile\end{tabular}        & $\checkmark$ & $\checkmark$ &   &   &   &                       &                        & 59.2& 54.5& 0.73\\ \hline
 \begin{tabular}[c]{@{}c@{}}Vocal.\end{tabular} &   &   &   &   &   &  \begin{tabular}[c]{@{}c@{}}$\checkmark$\\ (RGB)\end{tabular}                     & \multicolumn{1}{c|}{$\checkmark$} & 63.2& 51.3& 0.77\\ \hline
\end{tabular}

\end{table*}

Table~\ref{tab:Behavior_model_performance} summarizes the performance of our behavior detection models across various feature combinations on a held-out test set with no subject overlap. For the gaze-related behaviors, progressive addition of contextual information in the form of explicit object position and 3D gaze direction significantly improved gaze to face (AUROC score improved from 0.67 to 0.76) and gaze to object (AUROC score improved from 0.64 to 0.71) detection performance. Smile detection model relying on landmark and facial AUs performs classification with AUROC of 0.73. Our earlier experiments on vocalization detection using facial landmarks showed low recall (24.0\%). Using Mel spectrogram features, we obtain recall of 51.3\% and AUROC score of 0.77.

\subsection{ASD Risk Prediction from Ground Truth Behaviors}
\label{sec:asd_ground_truth}
We first evaluate the Stage 2 risk classifier using manually-coded, time-aligned behaviors as input, establishing an upper bound on performance given perfect behavior detection. Feature construction, classifier choices, class imbalance handling, and the cross-validation protocol follow Section~\ref{sec:risk_prediction}.

\begin{table}[]
\caption{Performance metrics (sensitivity, specificity, positive predictive value, AUROC, and accuracy) of Logistic Regression (LR), Random Forest (RF), and Multilayer Perceptron (MLP) models for ASD behavior classification on ground truth labels. Models were trained on a balanced dataset using SMOTE and Tomek Links to address the 93:7 Non-ASD to ASD class imbalance, and evaluated via stratified 10-fold cross-validation. \textbf{Values in parentheses represent the 95\% confidence interval.} These results use an earlier 10-fold CV protocol for classifier selection; the ground-truth baseline used in the subsequent component-validation, calibration, and within-visit partner-contrast analyses is the ground-truth row of Table~\ref{tab:ASD_classifier_ML}, computed under a new protocol.}

\label{tab:ASD_classifier_gt}
\centering
\small
\begin{tabular}{l *{5}{c}} 
\hline
    & \textbf{Sens (\%)} & \textbf{Spec (\%)} & \textbf{PPV (\%)} & \textbf{AUROC} & \textbf{Acc (\%)} \\
\hline
LR  & 69.3 & 86.0 & 29.1 & 0.81 & 84.4 \\
    & (64.3–74.3)& (83.5–88.5) &  (24.5–33.7) & (0.78–0.84) & (82.2–86.6) \\
\hline
RF  & 61.1 & 84.2 & \textbf{39.6} & 0.78 & 81.2 \\
    & (54.3–67.9) & (80.0–88.4) & (34.3–44.9) & (0.74-0.82) & (77.8–84.6) \\
\hline
MLP  & \textbf{76.0} & \textbf{85.9} & 31.2 & \textbf{0.81} & \textbf{85.1} \\
    &  (72.5–79.5) & (83.5–88.3) & (26.8–35.6) & (0.79–0.83) &  (82.9–87.3) \\
\hline
\end{tabular}
\end{table}

Table~\ref{tab:ASD_classifier_gt} shows the comparative performance of these classifiers. The MLP-based neural network yielded the most robust ASD risk prediction, achieving a sensitivity of 76.0\% and specificity of 85.9\%, with an overall AUROC of 0.81 and accuracy of 85.1\%. While the specific gold standard for screening instruments is often cited as 80\% sensitivity and specificity \citep{GOLDtrevethan2017sensitivity}, it is important to note that our prediction relies solely on social communication behavior codes observed during a brief 3-minute interactive play session. This constrained observation window stands in contrast to broader screening tools like M-CHAT or ITC, which assess a wider range of developmental domains over longer periods.

Crucially, the comparative analysis highlights significant divergences in model stability and sensitivity. As illustrated by the 95\% confidence intervals in Table~\ref{tab:ASD_classifier_gt}, the MLP demonstrated significantly tighter intervals across all metrics compared to the Random Forest (RF) and Logistic Regression (LR). This indicates that the MLP is less sensitive to the variations in data splits inherent in cross-validation. The performance gap was most reliable in sensitivity, the critical metric for screening. The RF model exhibited the highest variability and the lowest average sensitivity (61.1\%), suggesting it struggled to generalize the minority class features effectively. In contrast, the MLP achieved a distinct lead (76.0\%) while maintaining high specificity, striking the best balance between correctly identifying ASD behaviors and minimizing false positives. The MLP's advantage reflects its capacity to model non-linear interactions among co-occurring behavioral signals, which the axis-aligned splits of a tree-based model capture less naturally.

\begin{table*}[]
\caption{Component-validation sweep: ASD risk prediction when one or more ground-truth behavior channels are replaced by their ML-estimated counterparts, benchmarked against the all-ground-truth model. Sensitivity, specificity, PPV, AUROC, and accuracy are reported as mean over subject-disjoint stratified cross-validation folds (9 evaluation folds, 10-run ensemble per fold). \textbf{Values in parentheses are 95\% confidence intervals.} Single-channel substitutions are near-lossless; degradation is confined to the fully ML based configuration.}
\label{tab:ASD_classifier_ML}
\centering
\small
\renewcommand{\arraystretch}{1.3}
\begin{tabular}{l *{5}{c}}
\hline
                      & \textbf{Sens (\%)}   & \textbf{Spec (\%)}   & \textbf{PPV (\%)}   & \textbf{AUROC}      & \textbf{Acc (\%)}    \\
\hline
Look Face from ML                 & \makecell{76.3 \\ (67.7--84.9)} & \makecell{79.9 \\ (77.4--82.4)} & \makecell{23.5 \\ (19.8--27.2)} & \makecell{0.87 \\ (0.83--0.90)} & \makecell{79.7 \\ (77.2--82.1)} \\
Look Object from ML               & \makecell{80.9 \\ (75.0--86.9)} & \makecell{80.1 \\ (77.7--82.5)} & \makecell{24.7 \\ (21.8--27.5)} & \makecell{0.88 \\ (0.85--0.91)} & \makecell{80.2 \\ (78.1--82.3)} \\
Smile from ML                     & \makecell{80.7 \\ (74.6--86.9)} & \makecell{79.7 \\ (77.1--82.2)} & \makecell{24.4 \\ (21.5--27.4)} & \makecell{0.88 \\ (0.85--0.90)} & \makecell{79.8 \\ (77.5--82.0)} \\
Vocal from ML                     & \makecell{81.5 \\ (74.8--88.2)} & \makecell{79.8 \\ (76.9--82.7)} & \makecell{24.9 \\ (21.7--28.1)} & \makecell{0.87 \\ (0.84--0.90)} & \makecell{79.9 \\ (77.2--82.6)} \\
Look Face + Look Object           & \makecell{77.0 \\ (68.6--85.5)} & \makecell{79.6 \\ (77.1--82.1)} & \makecell{23.6 \\ (19.6--27.7)} & \makecell{0.85 \\ (0.81--0.89)} & \makecell{79.5 \\ (76.8--82.2)} \\
Look Face + Smile                 & \makecell{81.0 \\ (71.5--90.4)} & \makecell{78.4 \\ (76.0--80.9)} & \makecell{23.3 \\ (19.3--27.3)} & \makecell{0.86 \\ (0.83--0.90)} & \makecell{78.6 \\ (76.0--81.2)} \\
Look Face + Vocal                 & \makecell{74.5 \\ (66.2--82.9)} & \makecell{79.1 \\ (76.9--81.3)} & \makecell{22.5 \\ (19.2--25.7)} & \makecell{0.86 \\ (0.82--0.89)} & \makecell{78.7 \\ (76.5--80.9)} \\
Look Object + Smile               & \makecell{84.1 \\ (77.1--91.1)} & \makecell{78.1 \\ (75.0--81.1)} & \makecell{23.7 \\ (20.6--26.8)} & \makecell{0.87 \\ (0.84--0.90)} & \makecell{78.5 \\ (75.8--81.1)} \\
Look Object + Vocal               & \makecell{80.1 \\ (72.5--87.7)} & \makecell{79.1 \\ (75.9--82.3)} & \makecell{24.2 \\ (20.4--27.9)} & \makecell{0.87 \\ (0.83--0.90)} & \makecell{79.2 \\ (76.2--82.2)} \\
Smile + Vocal                     & \makecell{82.1 \\ (71.1--93.1)} & \makecell{77.2 \\ (74.2--80.2)} & \makecell{22.8 \\ (19.5--26.1)} & \makecell{0.87 \\ (0.84--0.89)} & \makecell{77.6 \\ (74.8--80.4)} \\
All behaviors from ML             & \makecell{69.4 \\ (60.7--78.1)} & \makecell{76.8 \\ (72.8--80.7)} & \makecell{19.8 \\ (16.9--22.8)} & \makecell{0.82 \\ (0.78--0.87)} & \makecell{76.2 \\ (72.8--79.6)} \\
\hline
\textbf{Ground Truth}             & \makecell{\textbf{80.9} \\ (75.2--86.6)} & \makecell{\textbf{81.3} \\ (79.0--83.6)} & \makecell{\textbf{26.0} \\ (22.7--29.3)} & \makecell{\textbf{0.88} \\ (0.86--0.90)} & \makecell{\textbf{81.3} \\ (79.2--83.4)} \\
\hline
\end{tabular}
\end{table*}

\subsection{ASD Risk Prediction from Computer Vision Assisted Behavior Estimation}
\label{subsec:ASD_pred_from_CV}

Having established the ground-truth upper bound, we next explore whether machine-estimated behaviors can stand in for human coding without degrading the risk classifier. To that end we replicate the Stage 2 experiment using machine-estimated behavioral features produced by the Stage 1 transformer models, following the leakage-preventing split and cross-validation protocol described in Section 4.3. Each behavior is substituted individually, then jointly, to isolate the contribution of each modality to the end-to-end performance gap.

\subsubsection{Component-level validation}

Table~\ref{tab:ASD_classifier_ML} reports the full substitution sweep. The central finding is that replacing any single ground-truth behavior with its ML estimate does not reduce accuracy and leads to almost identical values. AUROC moves from the all-ground-truth value of 0.88 to 0.87--0.88 for every individual channel, with sensitivity (76--82\%) and specificity ($\sim$80\%) overlapping the baseline confidence intervals. Gaze (``Look Face'', ``Look Object''), affect (``Smile''), and ``Vocal'' can individually substitute for manual codes with little loss in downstream classification performance, demonstrating their utility for automated behavioral analysis.

Degradation is confined to the fully automated ``All behaviors from ML'' configuration, where AUROC falls to 0.82 and PPV to 19.8\% against a ground-truth PPV of 26.0\%. This decline reflects a compounding of errors in the high-level social features: social smiling and social vocalization are contingent on accurate gaze-to-face detection, so errors in the underlying smile and vocalization models are exacerbated when gated by the gaze model. 

\subsection{Within-Visit Partner Contrast as a Discriminative Feature}
\label{sec:asymmetry_feature}

This section demonstrates a potential use case of ASD-FEAT's longitudinal, multiple-play-sessions-per-visit protocol: analysis of within-visit engagement across interaction partners, which the single-session datasets typical of automated ASD screening cannot support. The longitudinal ASD-FEAT dataset contains up to three sessions per subject in a single visit: one with a clinical examiner and up to two with the child's parent. This protocol enables a within-visit contrast between examiner-directed and parent-directed social behavior. Following the approach of \citet{Gangi2018gaze}, who compared infants' gaze to faces across familiar (parent) and unfamiliar (examiner) interaction partners, we quantify a parent-versus-examiner contrast for each behavior. Our analysis differs from that prior work in age range (18--36 months rather than infancy) and in diagnostic labeling (an ASD label assigned at each visit rather than a single 36-month outcome), and it extends the contrast beyond gaze to the affect and vocal channels. We treat this partner contrast as an engineered feature; following the convention of Sections \ref{sec:risk_prediction} and \ref{subsec:behavior_model_perf}--\ref{subsec:ASD_pred_from_CV}, we first quantify its per-visit discriminative power as a standalone signal, then evaluate it as an added input to the Stage~2 risk classifier. A detailed clinical interpretation of the observed per-channel patterns is left to future work.

\subsubsection{Within-visit partner contrast construction}

For each (subject, visit, behavior) we define a within-visit partner-contrast score:

\begin{equation}
c \;=\; \mu_{\text{Parent}} - \mu_{\text{Examiner}},
\qquad \mu_{\text{Parent}} = \text{mean}(\text{Parent}_1, \text{Parent}_2),
\label{eq:partner_contrast}
\end{equation}
where the partner-contrast score $c$ is the difference between the parent- and examiner-session means, and $\mu$ denotes the mean rate (or proportion) of the target behavior over the indicated session(s): $\mu_{\text{Parent}}$ averages the (up to two) parent sessions, while $\mu_{\text{Examiner}}$ is the value from the single examiner session. For each (behavior, age) cell, the effect size reported below is the between-group difference in the mean partner contrast,
\begin{equation}
\Delta \;=\; \bar{c}_{\text{ASD}} - \bar{c}_{\text{Non-ASD}},
\label{eq:delta}
\end{equation}
where $\bar{c}_{g}$ denotes the mean of $c$ over the subject-visits in diagnostic group $g$. The two aggregations are at different levels: $\mu$ in Eq.~\ref{eq:partner_contrast} averages sessions within a visit, whereas $\bar{c}$ in Eq.~\ref{eq:delta} averages the per-visit contrast across subjects within a diagnostic group. Positive contrast values indicate more of the target behavior in the parent session than the examiner session; negative values indicate the reverse; values near zero indicate balanced production across partners. The feature is motivated by a well-documented pattern: typically developing children are sensitive to partner identity and show a bias toward the familiar parent, whereas children with autism differentiate less between familiar and unfamiliar partners. Infants later diagnosed with autism look at parents and examiners at similar rates, while non-autistic infants look more at the parent~\citep{Gangi2018gaze}, and a reduced preference for familiar over unfamiliar faces has been reported both in eye-tracking of faces~\citep{sterling2008role} and in young children's neural responses to their mother's face~\citep{dawson2002neural}. A within-visit partner contrast should therefore be discriminative because it indexes this partner differentiation, which is attenuated in autism. We remain agnostic about the direction of the typical bias, which may differ across channels; the feature captures both directions through its sign, and the empirical pattern is reported per channel in the results.

We use rate and proportion forms of each behavior, defined as in Section~\ref{sec:risk_prediction}; the normalization controls for the uneven session lengths across the examiner and parent partners within a visit.

We use twelve features (six behaviors each in rate and proportion form) covering the gaze, affect, and vocal channels: Look Face, Look Object, Smile, Vocal, Social smile, and Social vocalization. Analysis is restricted to 18-, 24-, and 36-month visits, the ages at which subjects with an ASD label are present in non-trivial numbers, yielding 279 examiner-and-parent dual-session subject-visit instances from 118 subjects at these ages.

\subsubsection{Evaluation}

For each (behavior, age) combination, we treat the within-visit partner contrast as a one-dimensional ASD-risk score and fit an L2-regularized logistic regression as the per-cell classifier, evaluated under stratified 10-fold cross-validation with SMOTE+Tomek imbalance handling on the training fold of each split. The imbalance-handling and cross-validation conventions mirror those of the Stage~2 protocol in Section~\ref{sec:eval_protocol}. We report mean cross-validated AUROC with 95\% CIs, alongside the between-group effect size $\Delta$ (Eq.~\ref{eq:delta}), which expresses the effect in interpretable behavioral units. We report $\Delta$ as observed, without assuming a normative sign.

\subsubsection{Results}

Table \ref{tab:auroc_asymmetry} reports cross-validated AUROC and $\Delta$ on the twelve-behavior set. At 24 months, vocalization rate is the strongest single behavior (AUROC=0.85 [0.81, 0.88], $\Delta=-4.30$ events/min), producing the largest single-channel between-group contrast. Vocalization proportion is the next strongest at the same age (AUROC=0.78 [0.67, 0.88]) and remains discriminative at 36m (AUROC=0.66 [0.56, 0.76]). The directed sub-channel, social vocalization rate, is a related but weaker signal at 24m (0.69 [0.63, 0.76], $\Delta=-0.89$ events/min). The gaze channel (look-face rate) is moderately discriminative at both ages (0.67 [0.58, 0.77] at 24m, 0.65 [0.55, 0.75] at 36m), with absolute contrast $\Delta \approx +1.6$ events/min; look-object rate has comparable mean AUROC (0.65 at both ages) but its CI just falls short of the threshold 0.55. The proportion-family features clear the threshold at 36m (smile prop 0.66 [0.55, 0.78], social-smile prop 0.65 [0.56, 0.74], social-vocal prop 0.67 [0.56, 0.78]), capturing duration-weighted differences not visible from the rate variable.

At 18 months the partner-contrast signal is weak and not yet discriminative: no 18m cell exceeds the threshold (the strongest, raw vocalization rate, reaches only AUROC 0.72 with a wide CI under 9-fold estimation), and the raw partner contrasts are roughly half their 24-month magnitude: vocalization-rate $\Delta$ grows from $-2.06$ at 18m to $-4.30$ at 24m, and look-face-rate $\Delta$ from $+0.93$ to $+1.86$. The discriminative dual-partner contrast therefore emerges between 18 and 24 months rather than being present from 18m. Cells with AUROC below 0.5 (e.g.\ smile rate at 24m) are interpreted as null cells in which the classifier, under SMOTE+Tomek with $n_{\text{ASD}}=12$--$20$, selected an arbitrary decision direction in the synthetic-minority space, not as features that predict in the opposite direction. Per-cell AUROCs span 0.65--0.85 across the bolded cells (i.e. exceeding threshold 0.55); in absolute behavioral units, the examiner-vs-parent contrast at 24 months is approximately 4.3 events/min for raw vocalization, 1.6--1.9 events/min for the gaze behaviors, and 0.9 events/min for social vocalization.

\begin{table}[tbp]
\centering
\caption{Per-age cross-validated AUROC and raw effect magnitude ($\Delta = \bar{c}_{\text{ASD}} - \bar{c}_{\text{Non}}$ events/min, Eq.~\ref{eq:delta}, under \texttt{asd\_longitudinal}) of the within-visit partner contrast on the twelve-feature set. AUROC values are mean over stratified folds (10-fold at 24m and 36m); 95\% CIs in square brackets are derived from the fold distribution. Bold indicates AUROC 95\% CI lower bound $\geq 0.55$. The 18-month column uses 9-fold CV, as only $n_{\text{ASD}}=9$ dual-session instances are present at 18m; its AUROC estimates are correspondingly noisier (wider CIs) and no 18m cell clears the bolding threshold, but the raw effect magnitudes $\Delta$ are reported on the same footing as the later ages.}
\label{tab:auroc_asymmetry}
\small
\begin{tabular}{@{}llll@{}}
\toprule
Behavior & 18m AUROC ($\Delta$) & 24m AUROC ($\Delta$) & 36m AUROC ($\Delta$) \\ \midrule

smile rate & \begin{tabular}[c]{@{}l@{}} 0.33 ($-$0.06) \\ {[0.16--0.50]} \end{tabular} & \begin{tabular}[c]{@{}l@{}} 0.28 (+0.08) \\ {[0.15--0.41]} \end{tabular} & \begin{tabular}[c]{@{}l@{}} 0.59 (+0.23) \\ {[0.46--0.72]} \end{tabular} \\ \addlinespace

smile prop & \begin{tabular}[c]{@{}l@{}} 0.47 (+0.04) \\ {[0.21--0.73]} \end{tabular} & \begin{tabular}[c]{@{}l@{}} 0.65 (+0.10) \\ {[0.49--0.80]} \end{tabular} & \begin{tabular}[c]{@{}l@{}} \textbf{0.66} (+0.11) \\ \textbf{[0.55--0.78]} \end{tabular} \\ \addlinespace

lookface rate & \begin{tabular}[c]{@{}l@{}} 0.57 (+0.93) \\ {[0.36--0.78]} \end{tabular} & \begin{tabular}[c]{@{}l@{}} \textbf{0.67} (+1.86) \\ \textbf{[0.58--0.77]} \end{tabular} & \begin{tabular}[c]{@{}l@{}} \textbf{0.65} (+1.52) \\ \textbf{[0.55--0.75]} \end{tabular} \\ \addlinespace

lookface prop & \begin{tabular}[c]{@{}l@{}} 0.25 (+0.02) \\ {[0.13--0.38]} \end{tabular} & \begin{tabular}[c]{@{}l@{}} 0.60 (+0.05) \\ {[0.49--0.71]} \end{tabular} & \begin{tabular}[c]{@{}l@{}} 0.60 (+0.02) \\ {[0.49--0.71]} \end{tabular} \\ \addlinespace

lookobject rate & \begin{tabular}[c]{@{}l@{}} 0.56 (+0.78) \\ {[0.32--0.80]} \end{tabular} & \begin{tabular}[c]{@{}l@{}} 0.65 (+1.56) \\ {[0.53--0.76]} \end{tabular} & \begin{tabular}[c]{@{}l@{}} 0.65 (+1.55) \\ {[0.54--0.75]} \end{tabular} \\ \addlinespace

lookobject prop & \begin{tabular}[c]{@{}l@{}} 0.29 ($-$0.02) \\ {[0.15--0.43]} \end{tabular} & \begin{tabular}[c]{@{}l@{}} 0.59 ($-$0.03) \\ {[0.46--0.72]} \end{tabular} & \begin{tabular}[c]{@{}l@{}} 0.49 (+0.03) \\ {[0.35--0.63]} \end{tabular} \\ \addlinespace

vocal rate & \begin{tabular}[c]{@{}l@{}} 0.72 ($-$2.06) \\ {[0.54--0.89]} \end{tabular} & \begin{tabular}[c]{@{}l@{}} \textbf{0.85} ($-$4.30) \\ \textbf{[0.81--0.88]} \end{tabular} & \begin{tabular}[c]{@{}l@{}} 0.66 ($-$2.11) \\ {[0.47--0.85]} \end{tabular} \\ \addlinespace

vocal prop & \begin{tabular}[c]{@{}l@{}} 0.67 ($-$0.03) \\ {[0.43--0.91]} \end{tabular} & \begin{tabular}[c]{@{}l@{}} \textbf{0.78} ($-$0.07) \\ \textbf{[0.67--0.88]} \end{tabular} & \begin{tabular}[c]{@{}l@{}} \textbf{0.66} ($-$0.10) \\ \textbf{[0.56--0.76]} \end{tabular} \\ \addlinespace

social smile rate & \begin{tabular}[c]{@{}l@{}} 0.31 ($-$0.09) \\ {[0.15--0.46]} \end{tabular} & \begin{tabular}[c]{@{}l@{}} 0.40 (+0.26) \\ {[0.23--0.58]} \end{tabular} & \begin{tabular}[c]{@{}l@{}} 0.59 (+0.30) \\ {[0.51--0.66]} \end{tabular} \\ \addlinespace

social smile prop & \begin{tabular}[c]{@{}l@{}} 0.47 (+0.04) \\ {[0.21--0.73]} \end{tabular} & \begin{tabular}[c]{@{}l@{}} 0.63 (+0.11) \\ {[0.50--0.77]} \end{tabular} & \begin{tabular}[c]{@{}l@{}} \textbf{0.65} (+0.11) \\ \textbf{[0.56--0.74]} \end{tabular} \\ \addlinespace

social vocal rate & \begin{tabular}[c]{@{}l@{}} 0.61 ($-$0.76) \\ {[0.48--0.75]} \end{tabular} & \begin{tabular}[c]{@{}l@{}} \textbf{0.69} ($-$0.89) \\ \textbf{[0.63--0.76]} \end{tabular} & \begin{tabular}[c]{@{}l@{}} 0.59 ($-$1.07) \\ {[0.46--0.72]} \end{tabular} \\ \addlinespace

social vocal prop & \begin{tabular}[c]{@{}l@{}} 0.62 ($-$0.02) \\ {[0.43--0.80]} \end{tabular} & \begin{tabular}[c]{@{}l@{}} \textbf{0.65} ($-$0.01) \\ \textbf{[0.56--0.74]} \end{tabular} & \begin{tabular}[c]{@{}l@{}} \textbf{0.67} ($-$0.05) \\ \textbf{[0.56--0.78]} \end{tabular} \\
\bottomrule
\end{tabular}
\end{table}

The composition of $\Delta$ is shown in Table~\ref{tab:partner_stratified_means}, which reports the underlying ASD and Non-ASD parent- and examiner-session means. Reading across a row gives the within-group parent$-$examiner difference for each group, and $\Delta$ is the difference between the two.

\begin{table*}[tbp]
\centering
\small
\caption{Partner-stratified mean behavior rates (events/min for rate features; proportions for prop features) from ground-truth, expert-coded behaviors, by ASD label and age, in the matched dual-partner cohort ($n=188$ subject-visit instances at 24m/36m, 32 ASD). For each group, Parent and Examiner are the per-(subject,\,visit) session means and P$-$E their within-group difference. The final column, $\Delta = (\text{ASD }P{-}E) - (\text{Non-ASD }P{-}E) = \bar{c}_{\text{ASD}} - \bar{c}_{\text{Non-ASD}}$ (Eq.~\ref{eq:delta}), is the between-group difference, identical to the $\Delta$ reported in Table~\ref{tab:auroc_asymmetry}. All quantities are rounded independently to two decimals, so within-row differences may differ from the displayed $\Delta$ by up to 0.01.}
\label{tab:partner_stratified_means}
\setlength{\tabcolsep}{5pt}
\renewcommand{\arraystretch}{1.05}
\begin{tabular}{@{}ll ccc ccc c@{}}
\toprule
 &  & \multicolumn{3}{c}{ASD} & \multicolumn{3}{c}{Non-ASD} & \\
\cmidrule(lr){3-5}\cmidrule(lr){6-8}
Behavior & Age & Parent & Examiner & P$-$E & Parent & Examiner & P$-$E & $\Delta$ \\
\midrule
smile rate & 24m & 0.94 & 0.83 & $+0.11$ & 1.63 & 1.60 & $+0.03$ & $+0.08$ \\
 & 36m & 1.39 & 1.38 & $+0.01$ & 1.98 & 2.21 & $-0.23$ & $+0.23$ \\
\addlinespace
smile prop & 24m & 0.09 & 0.05 & $+0.04$ & 0.14 & 0.21 & $-0.07$ & $+0.10$ \\
 & 36m & 0.15 & 0.14 & $+0.01$ & 0.23 & 0.33 & $-0.10$ & $+0.11$ \\
\addlinespace
lookface rate & 24m & 1.94 & 2.17 & $-0.23$ & 3.85 & 5.93 & $-2.08$ & $+1.86$ \\
 & 36m & 2.96 & 3.17 & $-0.21$ & 4.49 & 6.23 & $-1.74$ & $+1.52$ \\
\addlinespace
lookface prop & 24m & 0.06 & 0.06 & $0.00$ & 0.11 & 0.16 & $-0.05$ & $+0.05$ \\
 & 36m & 0.07 & 0.09 & $-0.02$ & 0.11 & 0.14 & $-0.03$ & $+0.02$ \\
\addlinespace
lookobject rate & 24m & 2.39 & 2.92 & $-0.53$ & 4.14 & 6.23 & $-2.09$ & $+1.56$ \\
 & 36m & 3.51 & 3.62 & $-0.11$ & 4.85 & 6.51 & $-1.66$ & $+1.55$ \\
\addlinespace
lookobject prop & 24m & 0.90 & 0.89 & $+0.01$ & 0.84 & 0.80 & $+0.04$ & $-0.03$ \\
 & 36m & 0.88 & 0.84 & $+0.04$ & 0.83 & 0.82 & $+0.01$ & $+0.03$ \\
\addlinespace
vocal rate & 24m & 3.22 & 3.94 & $-0.72$ & 8.95 & 5.37 & $+3.58$ & $-4.30$ \\
 & 36m & 8.72 & 7.13 & $+1.59$ & 9.90 & 6.21 & $+3.69$ & $-2.11$ \\
\addlinespace
vocal prop & 24m & 0.10 & 0.10 & $0.00$ & 0.17 & 0.11 & $+0.06$ & $-0.07$ \\
 & 36m & 0.23 & 0.21 & $+0.02$ & 0.26 & 0.14 & $+0.12$ & $-0.10$ \\
\addlinespace
social smile rate & 24m & 0.57 & 0.50 & $+0.07$ & 1.03 & 1.22 & $-0.19$ & $+0.26$ \\
 & 36m & 0.81 & 0.85 & $-0.04$ & 1.28 & 1.62 & $-0.34$ & $+0.30$ \\
\addlinespace
social smile prop & 24m & 0.07 & 0.03 & $+0.04$ & 0.11 & 0.18 & $-0.07$ & $+0.11$ \\
 & 36m & 0.10 & 0.09 & $+0.01$ & 0.17 & 0.27 & $-0.10$ & $+0.11$ \\
\addlinespace
social vocal rate & 24m & 0.93 & 1.03 & $-0.10$ & 3.62 & 2.83 & $+0.79$ & $-0.89$ \\
 & 36m & 2.83 & 2.85 & $-0.02$ & 4.25 & 3.21 & $+1.04$ & $-1.07$ \\
\addlinespace
social vocal prop & 24m & 0.03 & 0.02 & $+0.01$ & 0.08 & 0.06 & $+0.02$ & $-0.01$ \\
 & 36m & 0.08 & 0.08 & $0.00$ & 0.13 & 0.08 & $+0.05$ & $-0.05$ \\
\bottomrule
\end{tabular}
\end{table*}

The impact of substituting ground truth per-behavior partner-contrast signal with ML generated predictions is shown in Table~\ref{tab:auroc_asymmetry_ml} which contrasts the ground-truth (G) per-cell AUROC and effect with the fully ML-estimated (FOSV) version, with both columns computed on the same matched cohort. Most channels preserve their discriminative pattern under substitution: lookface rate at 24m is unchanged (0.67 in both), and vocal rate at 24m which is the strongest single signal in GT (0.85), remains discriminative under ML (0.76, $\Delta = -2.34$ events/min) while losing tightness in its confidence interval. Notably, two cells gain discriminability under ML substitution: smile rate at 24m (GT 0.28 $\rightarrow$ FOSV 0.63) and social smile rate at 24m (0.40 $\rightarrow$ 0.63), where the ML-detected behavior carries a clearer between-group contrast than the corresponding human-coded rate. Look-object channels show the largest degradation, with look object rate at 36m dropping from 0.65 (GT) to 0.46 (FOSV).

\begin{table*}[tbp]
\centering
\caption{Per-behavior within-visit partner-contrast AUROC under ML substitution: ground truth (G, identical to Table~\ref{tab:auroc_asymmetry}) vs.\ fully ML-estimated (FOSV) across 18, 24, and 36 months. Each cell shows AUROC (raw effect $\Delta = \bar{c}_{\text{ASD}} - \bar{c}_{\text{NonASD}}$, Eq.~\ref{eq:delta}) over 95\% CI. Bold = AUROC 95\% CI lower bound $\geq 0.55$. The 18-month columns use reduced-fold CV (9-fold for G, 8-fold for FOSV, reflecting $n_{\text{ASD}}=9$ and $8$ dual-session instances respectively) and are correspondingly noisier. $\Delta$ for FOSV is measured in ML-detected behavior units, which may differ in scale from human-coded units (e.g., ML lookface events/min vs.\ GT lookface events/min); the comparison is qualitative across columns and quantitative within column.}
\label{tab:auroc_asymmetry_ml}
\small
\renewcommand{\arraystretch}{1.15}
\begin{tabular}{@{}lcccccc@{}}
\toprule
 & \multicolumn{3}{c}{Ground Truth (G)} & \multicolumn{3}{c}{All ML (FOSV)} \\
\cmidrule(lr){2-4}\cmidrule(lr){5-7}
Behavior & 18m & 24m & 36m & 18m & 24m & 36m \\
\midrule
smile rate & \makecell{0.33 ($-0.06$)\\{[}0.16, 0.50{]}} & \makecell{0.28 ($+0.08$)\\{[}0.15, 0.41{]}} & \makecell{0.59 ($+0.23$)\\{[}0.46, 0.72{]}} & \makecell{0.62 ($+0.38$)\\{[}0.42, 0.82{]}} & \makecell{\textbf{0.63} ($+0.58$)\\\textbf{[0.52, 0.75]}} & \makecell{0.57 ($+0.39$)\\{[}0.50, 0.64{]}} \\
smile prop & \makecell{0.47 ($+0.04$)\\{[}0.21, 0.73{]}} & \makecell{0.65 ($+0.10$)\\{[}0.49, 0.80{]}} & \makecell{\textbf{0.66} ($+0.11$)\\\textbf{[0.55, 0.78]}} & \makecell{0.54 ($+0.06$)\\{[}0.31, 0.76{]}} & \makecell{\textbf{0.67} ($+0.08$)\\\textbf{[0.54, 0.79]}} & \makecell{0.67 ($+0.10$)\\{[}0.51, 0.83{]}} \\
lookface rate & \makecell{0.57 ($+0.93$)\\{[}0.36, 0.78{]}} & \makecell{\textbf{0.67} ($+1.86$)\\\textbf{[0.58, 0.77]}} & \makecell{\textbf{0.65} ($+1.52$)\\\textbf{[0.55, 0.75]}} & \makecell{\textbf{0.75} ($-1.29$)\\\textbf{[0.57, 0.92]}} & \makecell{0.67 ($+1.32$)\\{[}0.52, 0.81{]}} & \makecell{0.60 ($+0.52$)\\{[}0.47, 0.72{]}} \\
lookface prop & \makecell{0.25 ($+0.02$)\\{[}0.13, 0.38{]}} & \makecell{0.60 ($+0.05$)\\{[}0.49, 0.71{]}} & \makecell{0.60 ($+0.02$)\\{[}0.49, 0.71{]}} & \makecell{0.57 ($+0.02$)\\{[}0.39, 0.75{]}} & \makecell{\textbf{0.68} ($+0.06$)\\\textbf{[0.61, 0.75]}} & \makecell{0.42 ($-0.01$)\\{[}0.32, 0.52{]}} \\
lookobject rate & \makecell{0.56 ($+0.78$)\\{[}0.32, 0.80{]}} & \makecell{0.65 ($+1.56$)\\{[}0.53, 0.76{]}} & \makecell{0.65 ($+1.55$)\\{[}0.54, 0.75{]}} & \makecell{0.58 ($+0.60$)\\{[}0.42, 0.74{]}} & \makecell{0.55 ($+0.54$)\\{[}0.39, 0.70{]}} & \makecell{0.46 ($-0.26$)\\{[}0.29, 0.63{]}} \\
lookobject prop & \makecell{0.29 ($-0.02$)\\{[}0.15, 0.43{]}} & \makecell{0.59 ($-0.03$)\\{[}0.46, 0.72{]}} & \makecell{0.49 ($+0.03$)\\{[}0.35, 0.63{]}} & \makecell{0.56 ($-0.04$)\\{[}0.43, 0.68{]}} & \makecell{0.54 ($-0.01$)\\{[}0.38, 0.70{]}} & \makecell{0.56 ($+0.04$)\\{[}0.39, 0.72{]}} \\
vocal rate & \makecell{0.72 ($-2.06$)\\{[}0.54, 0.89{]}} & \makecell{\textbf{0.85} ($-4.30$)\\\textbf{[0.81, 0.88]}} & \makecell{\textbf{0.66} ($-2.11$)\\\textbf{[0.47, 0.85]}} & \makecell{0.59 ($-0.81$)\\{[}0.36, 0.82{]}} & \makecell{0.76 ($-2.34$)\\{[}0.57, 0.95{]}} & \makecell{0.63 ($-0.82$)\\{[}0.50, 0.76{]}} \\
vocal prop & \makecell{0.67 ($-0.03$)\\{[}0.43, 0.91{]}} & \makecell{\textbf{0.78} ($-0.07$)\\\textbf{[0.67, 0.88]}} & \makecell{\textbf{0.66} ($-0.10$)\\\textbf{[0.56, 0.76]}} & \makecell{0.52 ($-0.01$)\\{[}0.34, 0.69{]}} & \makecell{0.77 ($-0.12$)\\{[}0.59, 0.95{]}} & \makecell{0.62 ($-0.08$)\\{[}0.48, 0.77{]}} \\
social smile rate & \makecell{0.31 ($-0.09$)\\{[}0.15, 0.46{]}} & \makecell{0.40 ($+0.26$)\\{[}0.23, 0.58{]}} & \makecell{0.59 ($+0.30$)\\{[}0.51, 0.66{]}} & \makecell{0.65 ($+0.50$)\\{[}0.44, 0.86{]}} & \makecell{\textbf{0.63} ($+0.55$)\\\textbf{[0.50, 0.75]}} & \makecell{0.57 ($+0.47$)\\{[}0.42, 0.72{]}} \\
social smile prop & \makecell{0.47 ($+0.04$)\\{[}0.21, 0.73{]}} & \makecell{0.63 ($+0.11$)\\{[}0.50, 0.77{]}} & \makecell{\textbf{0.65} ($+0.11$)\\\textbf{[0.56, 0.74]}} & \makecell{0.59 ($+0.07$)\\{[}0.38, 0.79{]}} & \makecell{\textbf{0.67} ($+0.08$)\\\textbf{[0.59, 0.75]}} & \makecell{\textbf{0.67} ($+0.09$)\\\textbf{[0.54, 0.79]}} \\
social vocal rate & \makecell{0.61 ($-0.76$)\\{[}0.48, 0.75{]}} & \makecell{\textbf{0.69} ($-0.89$)\\\textbf{[0.63, 0.76]}} & \makecell{0.59 ($-1.07$)\\{[}0.46, 0.72{]}} & \makecell{0.54 ($-0.51$)\\{[}0.44, 0.65{]}} & \makecell{0.62 ($-0.55$)\\{[}0.46, 0.78{]}} & \makecell{0.36 ($-0.21$)\\{[}0.18, 0.54{]}} \\
social vocal prop & \makecell{0.62 ($-0.02$)\\{[}0.43, 0.80{]}} & \makecell{\textbf{0.65} ($-0.01$)\\\textbf{[0.56, 0.74]}} & \makecell{\textbf{0.67} ($-0.05$)\\\textbf{[0.56, 0.78]}} & \makecell{0.52 ($-0.02$)\\{[}0.38, 0.65{]}} & \makecell{0.65 ($-0.06$)\\{[}0.50, 0.79{]}} & \makecell{0.60 ($-0.04$)\\{[}0.42, 0.77{]}} \\
\bottomrule
\end{tabular}
\end{table*}

\subsubsection{Discussion}
\label{Discussion}

The within-visit partner contrast carries discriminative signal at 24 and 36 months across gaze, affect, vocal, and social-vocal channels, though its per-cell AUROCs and wide confidence intervals mark it as a candidate feature rather than a standalone screener; at 18 months it is weak and not yet discriminative, and the 18m estimates rest on a reduced-fold CV ($n_{\text{ASD}}=9$) and should be read as indicative of effect direction rather than as resolved AUROCs.

The 118-subject partner-contrast cohort overrepresents longitudinal completers and ASD-positive subjects relative to the full ASD-FEAT cohort: included subjects attended a mean of 4.0 visits versus 2.0 for excluded subjects, and 18.6\% of included subjects ever carried an ASD label versus 1.2\% of excluded subjects.
Effect sizes reported here are therefore conditional on the protocol-completing subset. Partner-contrast features are released alongside the standard rate and proportion features; their value as classifier inputs is quantified next.

A detailed clinical interpretation of the specific per-channel patterns observed here, including their relation to prior findings on how children differentiate familiar from unfamiliar interaction partners~\citep{Gangi2018gaze}, is left to future work. In this paper we evaluate the partner contrast primarily as an engineered feature and quantify its contribution to automated screening.

\subsubsection{Within-visit partner contrast as a classifier feature}
\label{subsubsec:asymmetry_classifier}

We now fold the twelve within-visit partner-contrast features into the Stage~2 classifier. For fair comparison, both the baseline and the augmented model are trained and evaluated on the same matched cohort i.e. the 279 subject-visit instances (41 ASD, prevalence 14.7\%) from the 118 subjects with both an examiner and a parent session at the 18/24/36-month visits. The baseline uses the standard 26-dim feature vector whereas the augmented model adds the twelve partner-contrast features for a 38-dim input. Because the protocol records multiple sessions per subject-visit, predictions are formed by averaging the per-session softmax probabilities within each subject-visit before thresholding, yielding one decision per instance and avoiding any feature-space averaging. Table~\ref{tab:asymmetry_classifier} reports both models for ground truth and the ML substitutions of primary interest.

For the fully ML-estimated Look Face~+~Smile configuration, a two-channel combination spanning the gaze and affect modalities, augmentation raises sensitivity from 64.3\% to 75.6\% at comparable specificity (80.6\%~$\rightarrow$~82.8\%), with AUROC lifting from 0.81 to 0.85. Look Face~+~Vocal shows a similar pattern: sensitivity climbs from 50.9\% to 78.3\% at an 8-point specificity trade-off (86.5\%~$\rightarrow$~78.5\%), with AUROC rising from 0.78 to 0.84. The partner-contrast signal also partially rescues the fully automated ``All behaviors'' configuration, lifting sensitivity from 46.3\% to 50.6\%, specificity from 72.1\% to 77.3\%, and AUROC from 0.69 to 0.73. Every single- and two-channel ML configuration improves under augmentation: the augmented Look~Face (AUROC 0.85), Look~Object (0.84), Smile (0.86), and Vocal (0.83) all reach or surpass the 26-dim clean ground-truth baseline (AUROC 0.83). The same improvement appears for ground truth (AUROC 0.83~$\rightarrow$~0.85, sensitivity 63.3\%~$\rightarrow$~70.4\%), confirming that the parent-versus-examiner contrast carries discriminative within-visit information not present in aggregate single-partner rates. The corresponding prevalence-calibrated and imbalance-robust metrics are deferred to Section~\ref{subsec:pop_scale} and discussed alongside the full-cohort component-validation results.

\begin{table}[tbp]
\caption{Stage~2 risk prediction on the matched dual-partner cohort (279 subject-visit instances from 118 subjects, 41 ASD): standard 26-dim feature vector vs.\ 38-dim model augmented with the twelve within-visit partner-contrast features. Rows use ground-truth behaviors or the indicated ML substitution. Mean over subject-disjoint folds; 95\% CI ($\pm1.96\,\mathrm{SE}$) in parentheses. Bold marks Look Face~+~Smile and Look Face~+~Vocal, the two-channel ML configurations highlighted in the discussion (extended metrics in Table~\ref{tab:extended_metrics}, Panel~C).}
\label{tab:asymmetry_classifier}
\centering
\small
\renewcommand{\arraystretch}{1.2}
\begin{tabular}{l c *{5}{c}}
\toprule
 & \textbf{Dim} & \textbf{Sens (\%)} & \textbf{Spec (\%)} & \textbf{PPV (\%)} & \textbf{AUROC} & \textbf{Acc (\%)} \\
\midrule
Ground Truth & 26 & \makecell{63.3 \\ (47.8--78.9)} & \makecell{85.1 \\ (81.5--88.6)} & \makecell{42.1 \\ (33.1--51.1)} & \makecell{0.83 \\ (0.76--0.89)} & \makecell{82.1 \\ (78.3--85.9)} \\
Ground Truth & 38 & \makecell{70.4 \\ (60.5--80.2)} & \makecell{86.9 \\ (81.7--92.1)} & \makecell{49.4 \\ (40.6--58.3)} & \makecell{0.85 \\ (0.79--0.91)} & \makecell{84.5 \\ (80.4--88.6)} \\
\addlinespace
Look Face & 26 & \makecell{60.2 \\ (45.7--74.6)} & \makecell{82.8 \\ (77.5--88.0)} & \makecell{40.2 \\ (31.7--48.8)} & \makecell{0.79 \\ (0.74--0.84)} & \makecell{79.4 \\ (75.7--83.1)} \\
Look Face & 38 & \makecell{75.6 \\ (59.2--91.9)} & \makecell{80.0 \\ (74.4--85.6)} & \makecell{40.4 \\ (32.7--48.1)} & \makecell{0.85 \\ (0.80--0.89)} & \makecell{79.2 \\ (74.8--83.6)} \\
\addlinespace
Look Object & 26 & \makecell{63.3 \\ (47.8--78.9)} & \makecell{80.6 \\ (75.0--86.2)} & \makecell{37.0 \\ (28.7--45.2)} & \makecell{0.81 \\ (0.75--0.87)} & \makecell{78.2 \\ (73.8--82.5)} \\
Look Object & 38 & \makecell{73.3 \\ (56.7--90.0)} & \makecell{84.6 \\ (79.5--89.8)} & \makecell{45.0 \\ (37.1--52.9)} & \makecell{0.84 \\ (0.80--0.88)} & \makecell{82.8 \\ (78.7--86.9)} \\
\addlinespace
Smile & 26 & \makecell{60.0 \\ (45.4--74.6)} & \makecell{83.8 \\ (80.4--87.1)} & \makecell{38.7 \\ (30.2--47.3)} & \makecell{0.83 \\ (0.77--0.89)} & \makecell{80.6 \\ (77.1--84.0)} \\
Smile & 38 & \makecell{64.8 \\ (51.0--78.6)} & \makecell{84.2 \\ (78.4--90.1)} & \makecell{43.9 \\ (36.1--51.7)} & \makecell{0.86 \\ (0.82--0.90)} & \makecell{81.4 \\ (77.5--85.3)} \\
\addlinespace
Vocal & 26 & \makecell{57.4 \\ (43.3--71.5)} & \makecell{87.1 \\ (82.6--91.6)} & \makecell{45.5 \\ (33.7--57.3)} & \makecell{0.81 \\ (0.73--0.88)} & \makecell{82.6 \\ (78.0--87.1)} \\
Vocal & 38 & \makecell{61.1 \\ (50.2--72.0)} & \makecell{86.2 \\ (81.8--90.5)} & \makecell{45.2 \\ (35.1--55.3)} & \makecell{0.83 \\ (0.79--0.87)} & \makecell{82.6 \\ (79.0--86.2)} \\
\addlinespace
Look Face + Smile & 26 & \makecell{64.3 \\ (53.8--74.7)} & \makecell{80.6 \\ (77.2--84.0)} & \makecell{36.5 \\ (31.1--42.0)} & \makecell{0.81 \\ (0.76--0.87)} & \makecell{78.2 \\ (75.6--80.8)} \\
Look Face + Smile & 38 & \makecell{75.6 \\ (64.0--87.2)} & \makecell{82.8 \\ (77.9--87.8)} & \makecell{44.8 \\ (36.4--53.2)} & \makecell{0.85 \\ (0.81--0.89)} & \makecell{81.6 \\ (77.4--85.7)} \\
\addlinespace
Look Face + Vocal & 26 & \makecell{50.9 \\ (32.5--69.4)} & \makecell{86.5 \\ (81.1--92.0)} & \makecell{44.1 \\ (23.3--64.8)} & \makecell{0.78 \\ (0.71--0.86)} & \makecell{81.3 \\ (75.7--86.9)} \\
Look Face + Vocal & 38 & \makecell{78.3 \\ (63.2--93.5)} & \makecell{78.5 \\ (73.8--83.2)} & \makecell{39.3 \\ (32.5--46.1)} & \makecell{0.84 \\ (0.80--0.88)} & \makecell{78.3 \\ (75.0--81.7)} \\
\addlinespace
All behaviors & 26 & \makecell{46.3 \\ (29.4--63.2)} & \makecell{72.1 \\ (59.4--84.8)} & \makecell{24.8 \\ (17.0--32.6)} & \makecell{0.69 \\ (0.59--0.78)} & \makecell{68.8 \\ (58.9--78.7)} \\
All behaviors & 38 & \makecell{50.6 \\ (37.2--63.9)} & \makecell{77.3 \\ (69.6--84.9)} & \makecell{33.1 \\ (24.0--42.2)} & \makecell{0.73 \\ (0.69--0.78)} & \makecell{73.5 \\ (66.7--80.2)} \\
\bottomrule
\end{tabular}
\end{table}

\subsection{Population-scale clinical utility}
\label{subsec:pop_scale}

The cohort-level PPV values reported in Sections~\ref{subsec:ASD_pred_from_CV} and~\ref{sec:asymmetry_feature} are optimistic because both research cohorts are prevalence-enriched ($\sim$7.5\% ASD in the full Stage~2 cohort; 14.7\% in the matched dual-partner cohort) relative to the $\sim$3.2\% U.S. general-population rate. We recalibrate PPV to 3.2\% prevalence via Bayes' rule, following the adjusted-PPV reporting practice for ASD digital-phenotyping screeners~\citep{perochon2023early}. We also report two prevalence-independent diagnostic-utility indices long-standard in clinical epidemiology~\citep{jaeschke1994users}: the positive and negative likelihood ratios ($\mathrm{LR}^{+}=\mathrm{Sens}/(1-\mathrm{Spec})$, $\mathrm{LR}^{-}=(1-\mathrm{Sens})/\mathrm{Spec}$), and the Matthews correlation coefficient (MCC), which integrates all four confusion-matrix cells and avoids the accuracy bias that arises under severe class imbalance~\citep{chicco2020advantages}. Table~\ref{tab:extended_metrics} reports these metrics across all three experiments: Panel~A covers the full-cohort component-validation sweep (Section~\ref{subsec:ASD_pred_from_CV}); Panels~B and~C cover the matched-cohort baseline and partner-contrast-augmented models (Section~\ref{sec:asymmetry_feature}).

For the full-cohort component-validation sweep (Panel~A), the operating point of a single-session behavioral screener at population prevalence is a PPV near 12--13\%, not the $\sim$26\% suggested by the enriched cohort. Despite this gap, the ground-truth model attains $\mathrm{LR}^{+}=4.5$ and $\mathrm{LR}^{-}=0.23$, clinically credible values for a 3-minute play interaction, and single-channel ML substitution preserves them ($\mathrm{LR}^{+}=3.9$--$4.2$, MCC 0.35--0.38 vs.\ 0.39 ground truth). The fully automated model is the only configuration whose negative likelihood ratio degrades materially ($\mathrm{LR}^{-}=0.40$, MCC 0.28), indicating that component error accumulates as missed cases rather than uniform noise. These results set an upper bound on single-session screening from aggregate single-partner behavioral rates: whether human- or machine-coded, precision at population prevalence is limited by the low ASD base rate. This operating point warrants caution: at a 3.2\% base rate, a PPV of 12--13\% means that roughly seven of every eight positive screens would be false positives, so a single-session behavioral score is not adequate as a standalone screener. It sits at or below the range reported for the established M-CHAT/M-CHAT-R/F instruments, whose PPV runs from about 0.15 in real-world primary-care use without systematic follow-up~\citep{guthrie2019accuracy} up to 0.53--0.83 in research settings even before the confirmatory follow-up interview~\citep{yuen2018accuracy}, and higher still for the two-stage M-CHAT-R/F. Much of that advantage comes from a second confirmatory stage that our single-stage operating point lacks. We therefore read this result not as a deployable screener but as evidence that a brief automated interaction carries discriminative information, most plausibly as one input to a multi-stage screening or triage workflow rather than a replacement for validated instruments.

For the matched dual-partner cohort (Panels~B and~C), partner-contrast augmentation improves the calibrated metrics consistently across configurations. Among ML substitutions, augmented Look Face~+~Smile reaches MCC 0.488 with $\mathrm{LR}^{-}$ tightening from 0.44 to 0.29; Look Face~+~Vocal reaches MCC 0.450 with $\mathrm{LR}^{-}$ tightening from 0.58 to 0.26, both exceeding the unaugmented clean ground-truth baseline (MCC 0.415, $\mathrm{LR}^{-}=0.43$). The augmentation also partially rescues the fully automated ``All behaviors'' configuration (MCC 0.158\,$\rightarrow$\,0.254, $\mathrm{LR}^{-}=0.76\,\rightarrow\,0.64$), though it remains below the clean ground-truth baseline. The augmented ground-truth model itself (MCC 0.504, $\mathrm{LR}^{+}=6.5$, $\mathrm{LR}^{-}=0.34$) sets the cohort ceiling and lifts PPV at population prevalence from 13.5\% to 17.2\%. 
The partner contrast improves discrimination and recovers missed cases, but does not lift precision past the ceiling that the low ASD base rate imposes on any single-visit screen based on aggregate behavior rates.

\begin{table}[tbp]
\caption{Prevalence-calibrated and imbalance-robust metrics. PPV@3.2\% recalibrates positive predictive value to the U.S.\ general-population ASD prevalence via Bayes' rule (95\% CI in parentheses); $\mathrm{LR}^{+}$/$\mathrm{LR}^{-}$ are prevalence-independent likelihood ratios; MCC is the Matthews correlation coefficient evaluated at the cohort prevalence. Panel A: full Stage~2 cohort (Table~\ref{tab:ASD_classifier_ML} protocol). Panels B--C: matched dual-partner cohort, 26-dim baseline vs.\ 38-dim partner-contrast-augmented (Table~\ref{tab:asymmetry_classifier} protocol).}
\label{tab:extended_metrics}
\centering
\small
\begin{tabular}{@{}lcccc@{}}
\toprule
 & \textbf{MCC} & \textbf{LR$^{+}$} & \textbf{LR$^{-}$} & \textbf{PPV@3.2\% (\%)} \\
\midrule
\multicolumn{5}{@{}l}{\textit{Panel A: full cohort (component-validation sweep)}} \\
Ground Truth      & 0.389 & 4.45 & 0.23 & 12.8 (11.3--14.2) \\
Look Face         & 0.347 & 3.94 & 0.30 & 11.4 (9.6--13.3) \\
Look Object       & 0.376 & 4.18 & 0.24 & 12.1 (10.7--13.4) \\
Smile             & 0.370 & 4.07 & 0.24 & 11.8 (10.6--13.0) \\
Vocal             & 0.377 & 4.21 & 0.23 & 12.2 (10.5--13.9) \\
Look Face + Smile & 0.358 & 3.89 & 0.25 & 11.3 (9.4--13.2) \\
Look Face + Vocal & 0.328 & 3.66 & 0.32 & 10.7 (9.1--12.3) \\
All behaviors     & 0.277 & 3.08 & 0.40 & 9.2 (8.1--10.3) \\
\midrule
\multicolumn{5}{@{}l}{\textit{Panel B: matched cohort, 26-dim baseline}} \\
Ground Truth      & 0.415 & 4.93 & 0.43 & 13.5 (8.9--18.0) \\
Look Face         & 0.371 & 4.45 & 0.47 & 12.4 (8.5--16.3) \\
Look Object       & 0.361 & 3.62 & 0.45 & 10.5 (8.1--13.0) \\
Smile             & 0.370 & 3.94 & 0.48 & 11.4 (8.8--14.0) \\
Vocal             & 0.409 & 6.11 & 0.50 & 15.8 (9.6--21.9) \\
Look Face + Smile & 0.366 & 3.48 & 0.44 & 10.3 (8.6--11.9) \\
Look Face + Vocal & 0.354 & 4.91 & 0.58 & 22.2 (1.9--42.5) \\
All behaviors     & 0.158 & 2.11 & 0.76 & 6.4 (3.9--8.9) \\
\midrule
\multicolumn{5}{@{}l}{\textit{Panel C: matched cohort, 38-dim partner-contrast-augmented}} \\
Ground Truth      & 0.504 & 6.49 & 0.34 & 17.2 (12.8--21.5) \\
Look Face         & 0.450 & 4.86 & 0.30 & 13.2 (8.3--18.1) \\
Look Object       & 0.487 & 5.86 & 0.31 & 15.5 (10.4--20.7) \\
Smile             & 0.429 & 5.30 & 0.42 & 14.5 (10.4--18.6) \\
Vocal             & 0.425 & 5.75 & 0.45 & 15.2 (10.0--20.5) \\
\textbf{Look Face + Smile} & \textbf{0.488} & \textbf{5.67} & \textbf{0.29} & \textbf{15.2 (10.5--19.9)} \\
\textbf{Look Face + Vocal} & \textbf{0.450} & \textbf{4.29} & \textbf{0.26} & \textbf{12.1 (8.4--15.8)} \\
All behaviors     & 0.254 & 3.38 & 0.64 & 9.7 (5.9--13.5) \\
\bottomrule
\end{tabular}
\end{table}

\section{Conclusion}
We propose a two-stage ASD risk screening model that identifies clinically relevant social behaviors and uses their aggregate statistics as input features to perform ASD risk prediction. A central contribution is ASD-FEAT itself, a feature dataset derived from longitudinal infant--adult interaction videos. Our expert-guided, feature-based approach enables robust behavior detection in naturalistic, non-ideal scenarios such as non-frontal face poses, occlusions, and lack of camera calibration. Using computer vision to analyze children's behaviors from head movements and gaze patterns enables quantification of delays in orienting to dynamic stimuli. Such delays, subtle and difficult to detect even for expert observers, can be quantified by digital systems. This capability can support early autism screening and promote deeper insights into the nature of autistic behaviors. Automatic identification of autistic behaviors of interest offers potential benefits to clinicians by enabling faster and more consistent behavior coding, reducing manual coding effort, accelerating the overall screening process. Beyond aggregate behavioral rates, our analysis shows how this approach can surface potentially discriminative features within a dataset, such as the within-visit contrast between examiner-directed and parent-directed social behavior, that standard ground-truth coding and single-partner analyses do not easily capture. We also demonstrate that adding this within-visit partner contrast to the Stage~2 classifier lifts the fully automated Look Face~+~Smile and Look Face~+~Vocal configurations to Matthews correlations of 0.49 and 0.45 respectively, both exceeding the human-coded single-partner baseline of 0.42. This captures discriminative within-visit information across gaze, affect, vocal, and social-vocal channels that is not accessible in single-partner datasets.

\impact{
This dataset addresses key gaps in autism behavior research through privacy-preserving design by releasing derived features rather than raw video. At the same time, it maintains research utility through comprehensive multimodal representations and gold-standard diagnostic assessments across longitudinal timepoints. The approach demonstrates that interpretable behavioral features achieve competitive screening performance without specialized hardware, potentially expanding access in underserved communities facing specialist shortages and year-long diagnostic wait times. By lowering barriers to research participation, this resource may accelerate development of improved early detection tools during critical developmental windows when intervention is most effective.

\begin{sloppypar}
However, our predominantly U.S.-based sample limits generalizability. Behavioral norms for eye contact and social interaction vary substantially across cultures; deployment without rigorous cross-cultural validation risks systematic bias possibly resulting in over-identification of autism in populations where behaviors differ from training norms, under-identification in others. Additionally, our data was collected under a standardized recording setup (e.g., fixed camera position, child seated in a high chair); it remains unclear whether the extracted features and trained models will generalize to other configurations such as single-camera setups, floor-based play, or varied clinical environments. The enriched prevalence (17\% vs. 3.2\% population rate) requires careful recalibration for real-world contexts. Privacy risks persist despite feature-based design, and behavioral analysis technologies face dual-use concerns including non-consensual surveillance and discriminatory screening. Responsible impact requires transparent limitation reporting, multi-site validation prioritizing diverse populations, community engagement including autistic self-advocates, and acknowledgment that dataset composition places inherent limits on model generalization. We view this release as a step toward advancing responsible pediatric behavioral data sharing, rather than a complete solution.
\end{sloppypar}}

\acks{This work is supported by NIH R01 MH121344. Computing
support for this work was provided by Oracle Cloud credits
under Oracle for Research program. This work used Delta advanced computing and data resource at the University of Illinois Urbana-Champaign through allocation CIS230383 from the Advanced Cyberinfrastructure Coordination Ecosystem: Services \& Support (ACCESS) program, which is supported by U.S. National Science Foundation grants \#2138259, \#2138286, \#2138307, \#2137603, and \#2138296.}

\vskip 0.2in
\bibliography{sample}

\clearpage
\appendix
\section{Datasheet for ASD-FEAT}
\label{app:datasheet}

This datasheet follows the template proposed by \citet{gebru2021datasheets}. Where an answer duplicates content already presented in the main text, we cite the relevant section rather than restate it.

\subsection{Motivation}

\paragraph{For what purpose was the dataset created?}
ASD-FEAT was created to enable reproducible machine learning research on early autism spectrum disorder (ASD) risk prediction from short adult--child interactive play videos, while preserving the privacy of pediatric participants. Existing public ASD behavioral datasets are either small, lack longitudinal coverage of the critical 6--36 month developmental window, or release raw video that limits sharing and reuse. ASD-FEAT addresses these gaps by releasing high-quality derived features and frame-level expert behavior annotations in lieu of raw recordings.

\paragraph{Who created the dataset and on behalf of which entity?}
The derived feature dataset was curated and generated by an interdisciplinary team of computer-vision researchers and developmental clinicians at the University of Kentucky and the University of California, Davis; see the title page for authorship and affiliations.

\paragraph{Who funded the creation of the dataset?}
This work was supported by NIH R01 MH121344, with computing support from the Oracle for Research program and NSF-funded ACCESS/Delta resources; see the main-paper acknowledgements for the full funding statement.

\subsection{Composition}

\paragraph{What do the instances represent?}
Each instance corresponds to a single 3-minute adult--child interactive play session, recorded as part of a longitudinal infant sibling study. Within each session, the feature data consists of frame-level rows consisting of derived per-frame features and expert binary behavior labels.

\paragraph{How many instances are there in total?}
The dataset comprises 1{,}886 sessions from 318 unique children, distributed across six assessment timepoints (6, 9, 12, 18, 24, and 36 months); see Table~\ref{tab:dataset_demographic_spec}.

\paragraph{Does the dataset contain all possible instances or a sample?}
The dataset is a complete release of all sessions in the source longitudinal study shared with the team at University of Kentucky and that passed quality control (face detection, audio integrity, annotator agreement). The cohort itself is an enriched-prevalence sample, not a population sample (see ``identification of subpopulations'' below).

\paragraph{What data does each instance consist of?}
Each session contains: (i) per-frame derived features --- 68-point 2D facial landmarks, 56-point 2D eye landmarks, facial action units (17 intensity and 18 presence channels), 3D gaze direction from Gaze360, an adult-head and a task-object bounding box, I3D RGB and optical-flow embeddings, and a mel spectrogram of the session audio; (ii) frame-level expert binary labels for the four directly-coded behaviors (Look Face, Look Object, Smile, Vocalization), from which two derived composite labels (Social Smile, Social Vocalization, Table~\ref{tab:behavioral_coding}) are computed for the Stage~2 ASD-classifier stage; and (iii) de-identified subject metadata (child age in months at the visit and child sex). The clinical ASD outcome (ASD or Not ASD) is released both per visit (\texttt{asd\_longitudinal}) and as the final 36-month diagnosis (\texttt{asd\_36}), indexed by subject/session ID alongside the per-session feature folders. No direct identifiers, globally unique identifiers (GUIDs), dates of birth, or exact ages are released. Tables~\ref{tab:dataset_feat_spec} and~\ref{tab:dataset_labels_spec} enumerate the field names.

\paragraph{Is there a label or target associated with each instance?}
Yes. At the frame level, four directly-coded binary behavior labels are provided per frame; two additional composite labels (Social Smile, Social Vocalization) are derived from these and from the Look Face track at the Stage~2 ASD-classifier stage. At the session level, the best-estimate clinical ASD diagnosis at the time of the visit (based on ADOS score above the ASD cutoff and DSM-5 fulfillment) is the Stage~2 prediction target; both the per-visit outcome and the final 36-month diagnosis are released.

\paragraph{Is any information missing from individual instances?}
Frame-level features can be missing when face detection fails (e.g., severe occlusion, off-camera head movement); these frames are flagged via a \texttt{valid} indicator and a tracking confidence score. Raw video and audio are deliberately not released.

\paragraph{Are relationships between individual instances made explicit?}
Yes. Sessions are linked to subjects via a de-identified subject identifier, and to visits via the age-at-visit field, enabling longitudinal analyses while preserving anonymity.

\paragraph{Are there recommended data splits?}
For Stage 1 behavior detection, we recommend a subject-disjoint train/validation/test split. For Stage 2 risk prediction, we use stratified 10-fold cross-validation with stratification over gender, age, and ASD diagnosis. When evaluating fully automated end-to-end performance, a 10\% holdout for Stage 1 training followed by 10-fold cross-validation on the remaining 90\% is required to prevent leakage between the two stages (Section~\ref{sec:risk_prediction}).

\paragraph{Are there errors, sources of noise, or redundancies?}
Manual annotations were performed by trained coders with double-coding on 15\% of sessions; intraclass correlation coefficients exceeded 0.90 for the four primary behaviors. Tracking-derived features inherit noise from the underlying OpenFace and Gaze360 models (e.g., reduced reliability under non-frontal head pose).

\paragraph{Is the dataset self-contained or does it link to external resources?}
Self-contained for all features and labels. The Kinetics-pretrained I3D weights used to extract video embeddings are external; the dataset distributes the embeddings rather than requiring users to re-run the extractor.

\paragraph{Does the dataset contain data that might be considered confidential?}
The source video recordings constitute protected health information under U.S. HIPAA. ASD-FEAT releases only derived features that, by design, do not permit reconstruction of identifiable video or audio.

\paragraph{Does the dataset contain data that might be offensive, insulting, threatening, or anxiety-inducing?}
No. Sessions consist of routine interactive play between a child and a parent or examiner.

\paragraph{Does the dataset relate to people?}
Yes. All instances are recordings of pediatric participants and an interacting adult.

\paragraph{Does the dataset identify subpopulations?}
Yes. Each session is labeled with the child's age, gender, and clinical ASD status. The cohort over-represents ASD-positive cases (17\%) relative to the general population (~3.2\%) due to enriched sampling of high-risk infant siblings; downstream users must recalibrate when generalizing to general-population settings.

\paragraph{Is it possible to identify individuals from the dataset?}
The release does not include any direct identifiers (name, date of birth, globally unique identifier (GUID), location, raw face image, or voice); ages are given only as the study's nominal assessment ages. Re-identification from derived geometric and embedding features is, to our knowledge, infeasible with current techniques; the release terms explicitly prohibit any re-identification attempt.

\paragraph{Does the dataset contain sensitive data?}
Yes --- pediatric clinical assessments. The dataset is distributed via Zenodo, a publicly funded, FAIR-compliant repository providing open access to the scientific community, with access managed through the repository's approved procedures rather than by the study team; see Section~\ref{sec:licensing}.

\subsection{Collection Process}

\paragraph{How was the data acquired?}
Source videos were collected during in-person clinical visits at a fixed-camera, high-chair recording setup. ASD-FEAT features were derived from those recordings using the pipeline described in Section~\ref{feature_description}.

\paragraph{What mechanisms were used to collect the data?}
Source video was recorded with stationary cameras and audio with synchronized microphones during structured 3-minute play sessions. Frame-level behavior coding was performed offline by trained coders using behavioral annotation software (Noldus Observer). The released dataset consists of features derived from the video and behavior labels, not the source videos.

\paragraph{If the dataset is a sample, what was the sampling strategy?}
Subjects were enrolled into a longitudinal infant sibling study with two strata: high-risk infants (those with at least one older sibling meeting ASD criteria) and low-risk infants (no family history of ASD). Within each visit, sessions were retained subject to quality control.

\paragraph{Who was involved in the data collection and how were they compensated?}
Clinical examiners and behavior coders were employed by the originating research institute. Participants and their guardians received compensation per the institute's IRB-approved protocol; specific amounts are not itemized here.

\paragraph{Over what timeframe was the data collected?}
The source videos were recorded between 2009 and 2019 as part of the longitudinal infant sibling study. Exact recording dates for individual sessions are not released, to reduce participant re-identification risk.

\paragraph{Were any ethical review processes conducted?}
Yes. The source study and the secondary release of derived features were both approved by the originating institute's Institutional Review Board.

\paragraph{Did individuals consent to the collection and use of their data?}
Yes. Written informed consent was obtained from a parent or legal guardian of every participating child, including consent for secondary research use of derived features.

\paragraph{Was a data protection impact analysis conducted?}
The release strategy consists of distributing derived features rather than raw audio-visual recordings, with open access managed by the hosting repository rather than the study team, and was designed in consultation with the originating institute's IRB and privacy office to mitigate re-identification risk.

\subsection{Preprocessing, Cleaning, and Labeling}

\paragraph{Was any preprocessing/cleaning/labeling done?}
Yes. Raw video frames were processed through OpenFace 2.0 (landmarks, AUs, head pose), Gaze360 (3D gaze), an object detector (toy bounding boxes), and I3D (spatiotemporal embeddings). Audio was converted to mel spectrograms at 48\,kHz. Behavior labels were assigned by trained coders at the frame level. Section~\ref{feature_description} describes each step.

\paragraph{Was the ``raw'' data saved?}
The source video and audio are retained by the originating institute under restricted access for clinical and IRB compliance purposes, but are not part of the ASD-FEAT release.

\paragraph{Is the software used to preprocess the data available?}
All upstream tools (OpenFace 2.0, Gaze360, Kinetics model for I3D) are publicly available; see Table~\ref{tab:upstream_licenses} for license terms. Reference scripts to reproduce the experiments in Sections~\ref{sec:framework} and~\ref{sec:experiments} are released alongside the dataset.

\subsection{Uses}

\paragraph{Has the dataset been used for any tasks already?}
Yes. This paper demonstrates the utility of the dataset by training transformer-based models to automatically identify the target social behaviors from the released features, and then using those automatically identified behaviors as inputs to a downstream ASD risk classifier. Performance of the fully automated pipeline is benchmarked against an upper-bound classifier trained on the manually coded ground-truth behaviors, allowing the cost of automation to be quantified directly (Sections~\ref{sec:framework} and~\ref{sec:experiments}).

\paragraph{Is there a repository linking to papers using the dataset?}
A list of citing publications will be maintained on the project landing page; see Section~\ref{sec:licensing}.

\paragraph{What other tasks could the dataset be used for?}
Beyond ASD risk prediction, the multimodal frame-level features support research on: developmental trajectories of social attention, joint attention and gaze-following modeling, child--adult interaction analysis, mel-spectrogram--based pediatric vocalization analysis, longitudinal change-point detection, fairness analysis across age and gender subgroups, and methodological work on event detection from noisy probability streams.

\paragraph{Is there anything about the composition that might impact future uses?}
The cohort is predominantly U.S.-based, recorded under a fixed clinical setup, and over-represents ASD due to enriched sampling. Models trained on ASD-FEAT may not generalize to other camera configurations, cultural contexts, or general-population prevalence without recalibration. See the impact statement in the main text for a fuller discussion.

\paragraph{Are there tasks for which the dataset should not be used?}
ASD-FEAT must not be used for: (i) any attempt to re-identify participants; (ii) commercial diagnostic deployment without independent clinical validation and regulatory clearance; (iii) deployment in settings substantially different from the cohort (e.g., non-clinical surveillance, employment screening, insurance underwriting); or (iv) research outside the consent scope obtained from participants and their guardians.

\subsection{Distribution}

\paragraph{Will the dataset be distributed to third parties?}
Yes, via Zenodo, a publicly funded, FAIR-compliant repository that provides secure, open access to the scientific community.

\paragraph{How will it be distributed and when?}
Via Zenodo, a publicly funded, FAIR-compliant repository, with a persistent DOI and open access managed through the repository's approved procedures rather than by the study team. The release will be available upon publication of this paper.

\paragraph{Will the dataset be distributed under a license?}
Yes. The dataset is released under an open-access license requiring citation, prohibiting re-identification attempts, and restricting use to research consistent with participant consent; see Section~\ref{sec:licensing} for full terms. Upstream tool licenses summarized in Table~\ref{tab:upstream_licenses} also propagate to downstream users.

\paragraph{Have any third parties imposed IP-based restrictions?}
The non-commercial terms of OpenFace 2.0 and Gaze360 propagate to derived features; downstream commercial use requires separate licensing arrangements with those upstream projects.

\paragraph{Do any export controls or regulatory restrictions apply?}
ASD-FEAT contains no export-controlled content. The release follows U.S. HIPAA-derived privacy norms reflected in the release terms.

\subsection{Maintenance}

\paragraph{Who will be supporting/hosting/maintaining the dataset?}
The originating research institute; the corresponding author can be contacted at \texttt{sccheung@ieee.org}.

\paragraph{How can the owner/curator/manager of the dataset be contacted?}
A monitored mailing list address will be provided on the Zenodo landing page in the camera-ready version.

\paragraph{Is there an erratum?}
Errata, when issued, will be posted as patch versions on Zenodo with their own DOIs.

\paragraph{Will the dataset be updated?}
Yes. We anticipate corrections (patch versions), modality extensions, and additional age timepoints (minor versions). All versions will remain accessible via their original DOIs to ensure citation stability.

\paragraph{If others want to extend or build on the dataset, is there a mechanism?}
External contributions of derived features computed from the released data are welcome and will be linked from the project landing page. Contributions requiring access to the original raw video must be coordinated with the originating institute under their IRB protocol.

\end{document}